\pdfoutput=1

\documentclass[11pt]{article}

\usepackage[]{ACL2023}

\usepackage{times}
\usepackage{latexsym}

\usepackage[T1]{fontenc}

\usepackage[utf8]{inputenc}

\usepackage{microtype}

\usepackage{inconsolata}
\usepackage{times}
\usepackage{latexsym}

\usepackage[T1]{fontenc}
\usepackage[utf8]{inputenc}
\usepackage{amsmath}
\usepackage{algorithm}
\usepackage{algpseudocode}
\usepackage{subfig}
\usepackage{multirow} 

\usepackage{times}
\usepackage{latexsym}
\usepackage{graphicx}
\usepackage{enumitem}

\title{\textbf{D-CALM}: A Dynamic Clustering-based Active Learning Approach for Mitigating Bias}

\author{Sabit Hassan and Malihe Alikhani \\
  School of Computing and Information \\
  University of Pittsburgh, Pittsburgh, PA \\
  \texttt{\{sah259,malihe\}@pitt.edu}}

\begin{document}
\maketitle
\begin{abstract}
Despite recent advancements, NLP models continue to be vulnerable to bias. This bias often originates from the uneven distribution of real-world data and can propagate through the annotation process. Escalated integration of these models in our lives calls for methods to mitigate bias without overbearing annotation costs. While \textbf{active learning (AL)} has shown promise in training models with a small amount of annotated data, AL's reliance on the model's behavior for selective sampling can lead to an accumulation of unwanted bias rather than bias mitigation. However, infusing clustering with AL can overcome the bias issue of both AL and traditional annotation methods while exploiting AL's annotation efficiency. In this paper, we propose a novel adaptive clustering-based active learning algorithm, \textbf{D-CALM}, that dynamically adjusts clustering and annotation efforts in response to an estimated classifier error-rate. Experiments on eight datasets for a diverse set of text classification tasks, including emotion, hatespeech, dialog act, and book type detection, demonstrate that our proposed algorithm significantly outperforms baseline AL approaches with both pretrained transformers and traditional Support Vector Machines. \textbf{D-CALM} showcases robustness against different measures of information gain and, as evident from our analysis of label and error distribution, can significantly reduce unwanted model bias. 
\end{abstract}

\section{Introduction}
\label{intro}

\begin{figure}
    \centering
    \includegraphics[width=\linewidth]{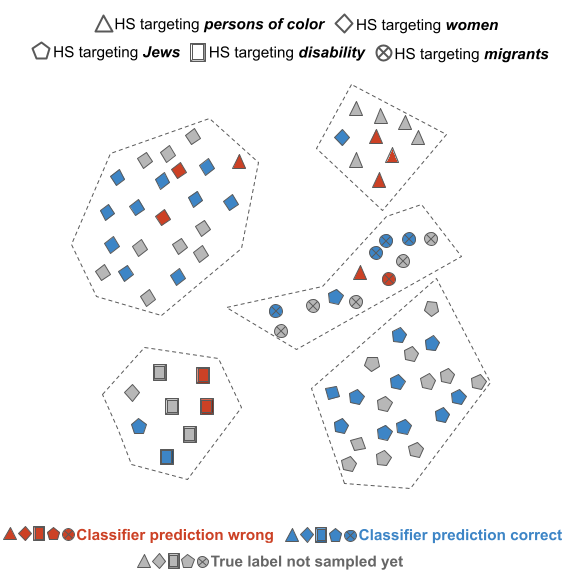}
    \caption{Example scenario: classifiers may not perform well for underrepresented groups in the data. Here, the classifier has a high error rate in detecting hatespeech (HS) against persons of color. Thus, annotation effort should be focused on regions (upper-right) more likely to contain hatespeech against persons of color.}
    \label{fig:intro_fig}
\end{figure}

While NLP models have experienced groundbreaking advancements in performance and functionality in recent years, they have been under scrutiny for exhibiting 
bias \cite{Lu2020GenderBI,ahn-oh-2021-mitigating,kiritchenko-mohammad-2018-examining}. As noted by \citet{davidson-etal-2019-racial}, classifier bias can stem from distribution in training data rather than the classifier itself. This bias is complex and can manifest in various forms, including racial, gender-based, and other types of discrimination. For example, in a hatespeech dataset, hatespeech against Persons of Color might be underrepresented, leading to a model biased against Persons of Color (Figure \ref{fig:intro_fig}). Since the true distribution of data is unknown prior to labeling, ridding these models of such unwanted bias would require annotating a large number of samples to ensure that minority groups are well-represented in the data, incurring much higher cost, time, and effort. As such, we are in need of methods that can mitigate unwanted bias without overwhelming annotation costs. We address the problem of bias with a novel \textit{clustering-based active learning} approach.  

Although active learning \cite{Settles2009ActiveLL} is regarded as an efficient method for training models, generic active learning methods can induce bias \cite{Krishnan2021MitigatingSB} rather than mitigate it. 
Although there have been numerous works aimed at mitigating bias of active learning methods by the machine learning community \cite{Farquhar2021OnSB,Gudovskiy2020DeepAL}, these approaches often necessitate an in-depth comprehension of machine learning and active learning theories. We hypothesize that infusing clustering with active learning will allow us to overcome bias issues of both generic active learning and traditional annotation approaches while leveraging the annotation efficiency of active learning. 



To this end, we propose a novel dynamic clustering-based algorithm that can substantially improve performance and mitigate bias \textemdash \textbf{\textbf{D-CALM}} (\textbf{D}ynamic \textbf{C}lustering-based \textbf{A}ctive \textbf{L}earning for \textbf{M}itigating Bias)\footnote{Our code is available at: \url{https://github.com/sabithsn/DCALM}}. \textbf{\textbf{D-CALM}} leverages the distance between a classifier's predictions and true labels in dynamically-adjusted subregions within the data. As opposed to existing active learning methods \cite{Bod2011ActiveLW,Berardo2015ActiveLW} that utilize static clustering of data, our proposed algorithm adapts the clustering in each iteration of active learning. As the classifier gets updated in each iteration, the classifier's error rate changes in different regions. By calibrating the boundaries of clusters iteratively, \textbf{\textbf{D-CALM}} focuses annotation effort in updated regions with the evolving classifier's error-rate. As \textbf{D-CALM} dynamically adapts its regions for obtaining samples, we hypothesize that our approach will result in reduced bias. Similar to \citet{hassan2018interactive}, we expect bias reduction to be reflected in improved performance metrics and more balanced label and error distribution. We test our hypothesis across eight datasets, spanning a diverse range of text classification tasks (e.g., fine-grained hatespeech, dialog act, emotion detection) and a case study of fine-grained hatespeech detection. Our algorithm is model agnostic, showing substantial improvement for both pretrained models and lightweight Support Vector Machines. Our experiments also demonstrate robustness of \textbf{\textbf{D-CALM}} with respect to different measures of information gain. 





\section{Related Work}
\label{related}

Active learning is a well-studied problem in machine learning \cite{Settles2009ActiveLL} with numerous scenarios and query strategies (Section 3). Although active learning has shown promise in many tasks, susceptibility to bias, particularly for neural networks, is a concern raised by several works \cite{yuan-etal-2020-cold}. There are existing works that aim to mitigate this bias. \citet{Farquhar2021OnSB} proposes using corrective weights to mitigate bias. \citet{Gudovskiy2020DeepAL} propose self-supervised Fischer-Kernel for active learning on biased datasets. These approaches, however, often require a deep understanding of active learning and neural networks. Our approach is tailored for the NLP community and can easily be deployed. 

In recent years, there has been a renewed interest in active learning within the NLP community \cite{zhang-etal-2022-survey}. Some recent works have applied active learning with BERT models for specific tasks such as intent classification \cite{Zhang2019ensemble}, sentence matching \cite{bai-etal-2020-pre}, parts-of-speech tagging \cite{chaudhary-etal-2021-reducing} or named entity recognition \cite{Liu2022LTPAN}. \citet{margatina-etal-2022-importance} propose continued pretraining on unlabeled data for active learning. \citet{rotman-reichart-2022-multi} adapt active learning to multi-task scenarios for transformer models. \citet{ein-dor-etal-2020-active} perform a large-scale empirical study of existing active learning strategies on binary classification tasks. In comparison, we target a diverse range of binary and multi-class classification tasks. 

Some other works in the NLP domain have adapted advanced active learning approaches. \citet{yuan-etal-2020-cold} adapt the BADGE \cite{Ash2020DeepBA} framework for active learning with BERT. While BADGE computes gradient embedding using the output layer of a neural network and then clusters the gradient space, \citet{yuan-etal-2020-cold} computes surprisal embeddings by using Masked Language Model loss. \citet{margatina-etal-2021-active} use acquisition functions to obtain contrastive samples for BERT. Our algorithm is comparatively straightforward, not requiring in-depth understanding of surprisal embeddings or acquisition functions. Our algorithm is also model-agnostic and can be applied to neural networks such as BERT, and traditional models such as SVMs. In addition, our clustering step relies on feature representation independent from the learner's representation which may induce bias during the learning process. While some of the aforementioned works \cite{ein-dor-etal-2020-active,yuan-etal-2020-cold,margatina-etal-2021-active} compute diversity in selected samples, our work is the first to analyze and address bias in active learning from a socio-cultural perspective.



\section{Background}
\label{framework}
This section presents the relevant background of generic active learning followed by a discussion of adapting clustering-based active learning framework for text classification. Within the scope of this paper, we focus on creating the train data. We assume that the dev and test data are already created. Literature of active \textit{testing} \cite{Kumar2018ClassifierRE,hassan2018interactive} can be referred to for efficiently creating the dev and test set.
\begin{figure}[!t]
    \centering
    \includegraphics[width=\linewidth]{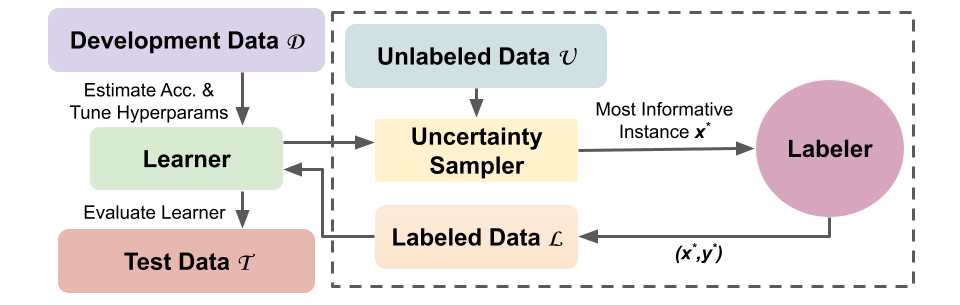}
    \caption{Active learning framework in a pool-based setting. Most informative sample from an unlabeled pool of data is annotated and added to training data.}
    \label{fig:al-framework}
\end{figure}

\subsection{Active Learning Framework}
Due to the expanse of active learning literature, it is important to define the generic active learning framework within the scope of this paper. To do so, we need to define the \textit{labeling scenario} and \textit{query-strategy}.


\subsubsection{Labeling Scenario} 

In our work, we assume there is a large pool of unlabeled dataset \textit{U} but only a small set of labeled dataset \textit{L} that can be obtained. \textit{L} is iteratively constructed by querying label for the \textit{most-informative} instance. We focus on \textit{pool-based} active learning because of its relevance to many recent NLP tasks (e.g., hatespeech detetion), for which, a large amount of unlabeled data is scraped from the web and then a subset of it is manually annotated.

\subsubsection{Query-Strategy}
Many types of query-strategies have been proposed for active learning over the years, including, but not limited to: uncertainty sampling \cite{Lewis1994ASA}, expected model change \cite{Settles2007MultipleInstanceAL}, expected error reduction \cite{Roy2001TowardOA}, and variance reduction \cite{Hoi2006LargescaleTC}. In our work, we focus on uncertainty sampling because of its popularity and synergy with pool-based sampling \cite{Settles2009ActiveLL}. \citet{Settles2009ActiveLL} lists three measures of uncertainty to identify most informative sample:
\begin{figure}[h]
    \centering
    \includegraphics[width=\linewidth]{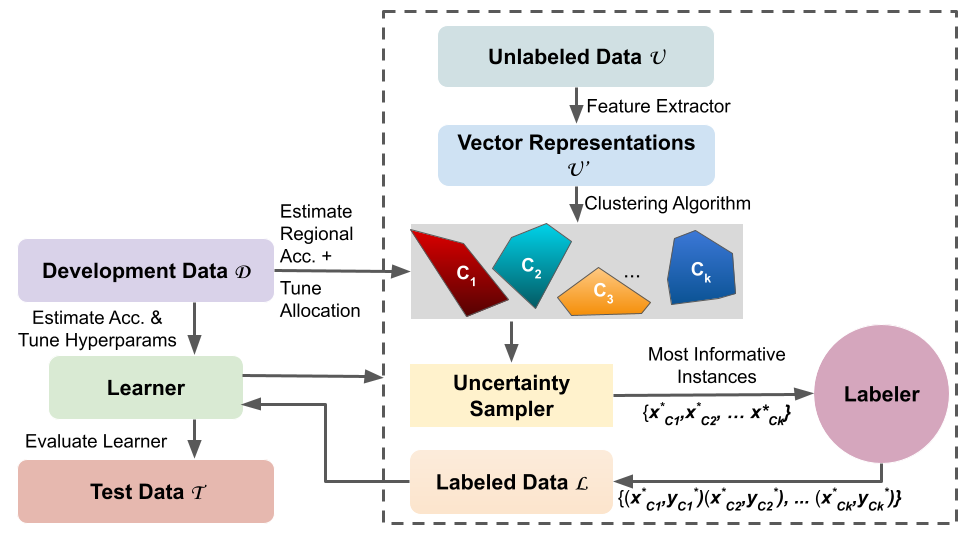}
    \caption{Clustering-based framework. First, unlabeled data is clustered and then most informative samples are chosen from each cluster.}
    \label{fig:cluster-framework}
\end{figure}
\paragraph{Least Confident:} Query the instance whose prediction is the least confident. 
\begin{equation}
\label{LC}
    x_{LC}^* = \underset{x}{argmax} \; 1 - P_\theta(\hat{y}|x)
\end{equation}
In Eq. \ref{LC}, $\hat{y} = argmax_yP_\theta(y|x)$, or the class label with the highest probability.

\paragraph{Smallest Margin:} Query the sample with minimum difference between two most likely classes:
\begin{equation}
\label{MS}
    x_{MS}^* = \underset{x}{argmin} \; P_\theta(\hat{y}_1|x) -P_\theta(\hat{y}_2|x)
\end{equation}

\paragraph{Entropy:} The most commonly used measure of uncertainty is entropy:
\begin{equation}
\label{Entropy}
    x_{E}^* = \underset{x}{argmax} \; - \underset{i}{\sum} P_\theta(y_i|x)logP_\theta(y_i|x)
\end{equation}
In Eq. \ref{Entropy}, i ranges over all possible labels.

It should be noted that, in binary classification, all the above measures become equivalent. 
The active learning framework, within the scope of this paper, is summarized in Figure \ref{fig:al-framework}

\subsubsection{Challenges}
\paragraph{Bias Induction:} Since the active learning framework relies on the model's uncertainty to choose samples, the framework may never query samples that the model is confident on. The active learning classifier can become \textit{confidently wrong} on certain samples, leading to an accumulation of bias.

\paragraph{Effective Batch Selection:} In a real-world setting, it is not feasible to obtain annotations one by one and queries need to be done in batches. The most straightforward approach would be to choose the \textit{\textbf{N}} most informative samples \cite{Citovsky2021BatchAL}. The limitations of this approach can be easily seen. Particularly when \textit{\textbf{N}} is large, it can amplify the bias discussed earlier.

\subsection{Clustering-based Framework} 

To address the challenges outlined earlier, we approach the problem with clustering-based framework for active learning under pool-based uncertainty sampling settings.
Within this framework, the first step is to obtain vector representation of the unlabeled data. This can be done using SentenceBERT \cite{reimers-gurevych-2019-sentence} or more traditional Doc2Vec \cite{Le2014DistributedRO}.

The next step is to cluster the data. This can be done using any clustering algorithm such as KMeans. Then, informative samples are chosen from each cluster, and are added to the training data. The classifier is retrained after each round and the process is repeated until the annotation budget runs out. Figure \ref{fig:cluster-framework} summarizes this framework.

\subsection{D-CALM}
\label{d-calm-algo}
Within the clustering-based framework of active learning, we propose a novel algorithm, \textbf{\textbf{D-CALM}}, that dynamically adjusts clusters in the data based on estimated classifier error rate.
\begin{algorithm}
\caption{\textbf{D-CALM}: \textbf{D}ynamic \textbf{C}lustering-based \textbf{A}ctive \textbf{L}earning for \textbf{M}itigating Bias}\label{alg:cap}
\begin{algorithmic}
\State $D,T \gets$ dev data, test data
\State $U,L \gets$ unlabeled data, labeled data
\State $G \gets$ bootstrapped classifier
\State $B \gets$ labeling budget
\State $N \gets$ annotation batch size
\State $m \gets$ initial number of clusters

\State Cluster $U$ into \{$C_1$, $C_2$, ... $C_m$\}
\State Partition $D$ into \{$C'_1$, $C'_2$, ... $C'_m$\}

\While{$B \geq 0$}
    \For{\texttt{i=0,1,...m}}
        \State Estimate accuracy $A_i$ in $C'_i$
    \EndFor

    \For{\texttt{i=0,1,...m}}
        \State Allocate $l_i = N*\frac{1-A_i}{\underset{j}{\sum}(1-A_j)}$
        \State Cluster $C_i$ into $\{C_{i_1}, C_{i_2}, ... C_{i_{l_i}}\}$
            \For{\texttt{j=0,1,...$l_i$}}
                \State $x_{ij}^* \gets$ most infor. sample in $C_{ij}$
                \State $y_{ij}^* \gets$ query true label for $x_{ij}^*$
                \State Add $(x_{ij}^*, y_{ij}^*)$ to $L$
            \EndFor
    \EndFor
    \State $G \gets$ retrain on $L$
    \State $B=B-N$
\EndWhile
\State Evaluate G on T
\end{algorithmic}
\end{algorithm}



 In our proposed algorithm, the cluster $C'_i$ is used to dynamically partition $C_i$. Our algorithm first observes how the classifier behaves in $C'_i$. For cluster $C_i$, it allocates samples proportional to the error rate in $C'_i$. Then the cluster $C_i$ is split into subclusters according to the number of samples allocated to $C_i$. Most informative sample from each subcluster is then added to training data. The subclusters are dynamically updated in each iteration to account for the classifier's new state. This prevents the classifier from repeatedly sampling from any particular region. It is worth noting that \textit{error-rate} can be substituted with different metrics to account for specific needs. For example, in scenarios where it is more important to reduce false negative rate reduce compared to false positive rate, the error rate can be substituted with false negative rate. In this paper, we focus on the general case of error-rate.  

\section{Experiment Setup}
In this section, we outline our experimental setup. 
\subsection{Active Learning Approaches}
For all the following approaches, total number of samples range from 100-300, initial allocation for bootstrapping is set to 50, and annotation batch size is 50. Similar to \cite{ein-dor-etal-2020-active}, the classifiers are retrained in each round. 
\paragraph{Random:} The allocated number of samples are picked randomly from the unlabeled pool.
\paragraph{TopN:} The classifier is bootstrapped with 50 samples. In each iteration N most informative samples are labeled and added to training data until labeling budget runs out. TopN is a widely used baseline \cite{yuan-etal-2020-cold,Ash2020DeepBA}.
\paragraph{Cluster-TopN:} The classifier is bootstrapped in the same way. The unlabeled pool is first clustered into 10 clusters and in each iteration, N/10 most informative samples are chosen from each cluster. Cluster-TopN combines TopN and stratified sampling \cite{qian-zhou-2010-clustering}. We choose Cluster-TopN as a baseline due to its similarity with multiple existing methods \cite{Xu2003RepresentativeSF,zhdanov2019diverse}. 
\paragraph{\textbf{D-CALM}:} The classifier is bootstrapped in a similar fashion. While \textbf{D-CALM} is not sensitive to the initial number of clusters because of its dynamic splitting into subclusters, we set initial number of clusters to 10 to be consistent with Cluster-TopN. 

\subsection{Models} 

\paragraph{Transformers} We fine-tune the widely-used bert-based-cased \cite{Devlin2019BERTPO}. We observed that the models stabilize on the dev data when fine-tuned for 5 epochs with learning rate of 8e-5 and batch size of 16. The same setting is used across all experiments.
\paragraph{Support Vector Machine (SVM)} We choose SVM as our alternate model as it is completely different from transformers and because SVMs are still in use for practical purposes due to speed and lightweight properties \cite{hassan-etal-2021-asad,hassan-etal-2022-cross}. We use Tf-IDF weighted character [2-5] grams to train SVMs with default scikit-learn settings\footnote{\url{https://scikit-learn.org/stable/modules/generated/sklearn.svm.LinearSVC.html}}. 

\subsection{Datasets}
We evaluate our proposed algorithm on eight diverse datasets, among which two are binary classification datasets and the rest are multiclass. 

\paragraph{BOOK32} \cite{iwana2016judging} contains 207K book titles categorized into 32 classes such as \textit{Biographies \& Memoirs}. We take a subset that contains 20K random samples from 10 most frequent classes for runtime efficiency. Random sampling ensures the subset respects original distribution. 

\paragraph{CONAN}\cite{fanton-etal-2021-human} contains 5K instances annotated for hatespeech targets: \textit{\textit{Disabled}, \textit{Jews}, \textit{LGBT+}, \textit{Migrants}, \textit{Muslims}, \textit{Person of Color (POC)}, \textit{Women}, and \textit{Other}}.

\paragraph{CARER} \cite{saravia-etal-2018-carer} is an emotion detection dataset that contains six basic emotions in the released version: \textit{Anger}, \textit{Fear}, \textit{Joy}, \textit{Love}, \textit{Sadness}, and \textit{Surprise}. \footnote{https://huggingface.co/datasets/emotion}. The released version consists of 16K training, 2K dev and 2K test instances. 

\paragraph{CoLA} \cite{saravia-etal-2018-carer} contains 9.5K sentences expertly annotated for acceptability (grammaticality) in the public version. We use the in-domain set as dev and out-of-domain as test set.

\paragraph{HATE} \cite{hateoffensive} contains a total of 24.7K tweets that are annotated as: \textit{Offensive}, \textit{Hatespeech}, and \textit{Neither}.

\paragraph{MRDA} \cite{shriberg-etal-2004-icsi} contains 117K instances annotated for dialog acts. We consider the five basic labels: \textit{Statement}, \textit{BackChannel}, \textit{Disruption}, \textit{FloorGrabber}, and \textit{Question}. We limit the data to 20K randomly chosen samples for runtime efficiency.

\paragraph{Q-Type} \cite{Li2002LearningQC} contains 5.5K train and 0.5K test instances annotated for question types. We take the first level of annotation, containing six classes: \textit{Entity, Description, Abbrebivation, Number, Human,} and textit{Location}.   

\paragraph{Subjectivity} \cite{Pang+Lee:04a} contains 10K snippets from Rotten Tomatoes/IMDB reviews automatically tagged as \textit{Subjective} or \textit{Objective}.

\subsection{Data Preparation}
\paragraph{Splits} 
We use default train-dev-test splits if they are provided. If they are not provided, we split the data into 70-10-20 splits. The train data is treated as unlabeled pool of data, dev data is used for tuning purposes and test data is used to report results. Table \ref{tab:stats} shows summary of data used.

\begin{table}[h]
    \centering
    \begin{tabular}{l|c|c|c|c}
    \hline
    \textbf{Dataset} & \textbf{classes} & \textbf{Pool} & \textbf{Dev} & \textbf{Test}\\
    \hline
    BOOK32 & 32 & 14K & 2K & 4K\\
    CONAN & 8 & 3.5K & 0.5K & 1K\\
    CARER & 6 & 16K & 2K & 4K\\
    CoLA & 2 & 8.5K & 0.5K & 0.5K\\
    Hatespeech & 3 & 17.2K & 2.4K & 4.9K\\
    MRDA & 5 & 14K & 2K & 4K\\
    Q-Type & 6 & 4.9K & 0.5K & 0.5K\\
    Subjectivity & 2 & 7K & 1K & 2K\\
    \hline
    \end{tabular}
    \caption{Statistics of used datasets}
    \label{tab:stats}
\end{table}

\paragraph{Vector Representation} We use MiniLM \cite{wang2020minilm} sentence-transformer to transform text instances into 384 dimensional vectors. These vectors are then used to cluster the unlabeled data.

\paragraph{Clustering}We use KMeans to cluster the unlabeled pool of data. We use scikit-learn\footnote{\url{https://scikit-learn.org/stable/modules/generated/sklearn.cluster.KMeans.html}} implementation of KMeans with default parameters.

\begin{figure*}[h]
\begin{centering}
\begin{subfloat}
  {\includegraphics[width=.316\textwidth,keepaspectratio]{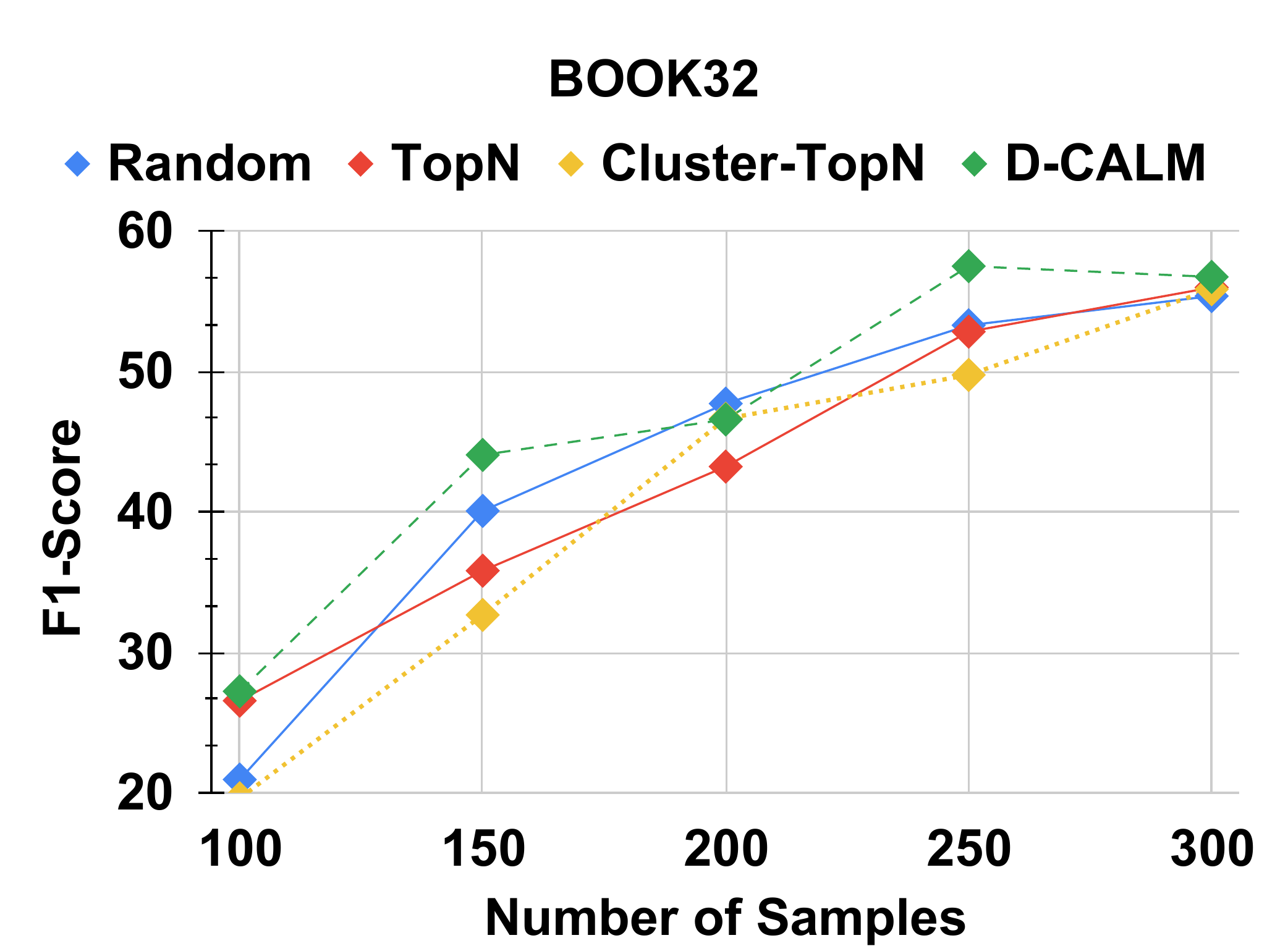}}
  \label{fig:sfig1b}
\end{subfloat}%
\begin{subfloat}
  {\includegraphics[width=.316\textwidth,,keepaspectratio]{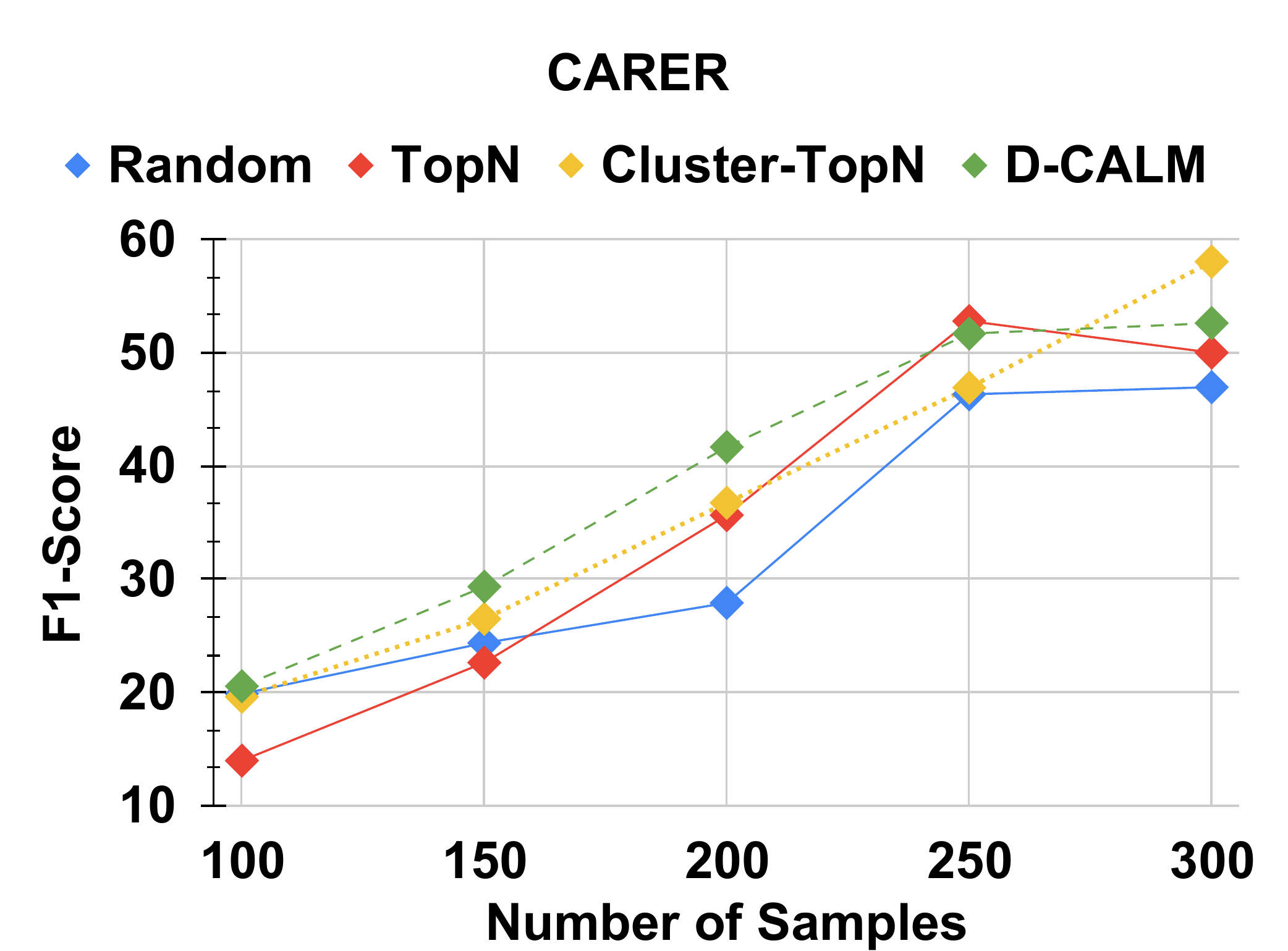}}
  \label{fig:sfig1c}
\end{subfloat}
\begin{subfloat}
  {\includegraphics[width=.316\textwidth,keepaspectratio]{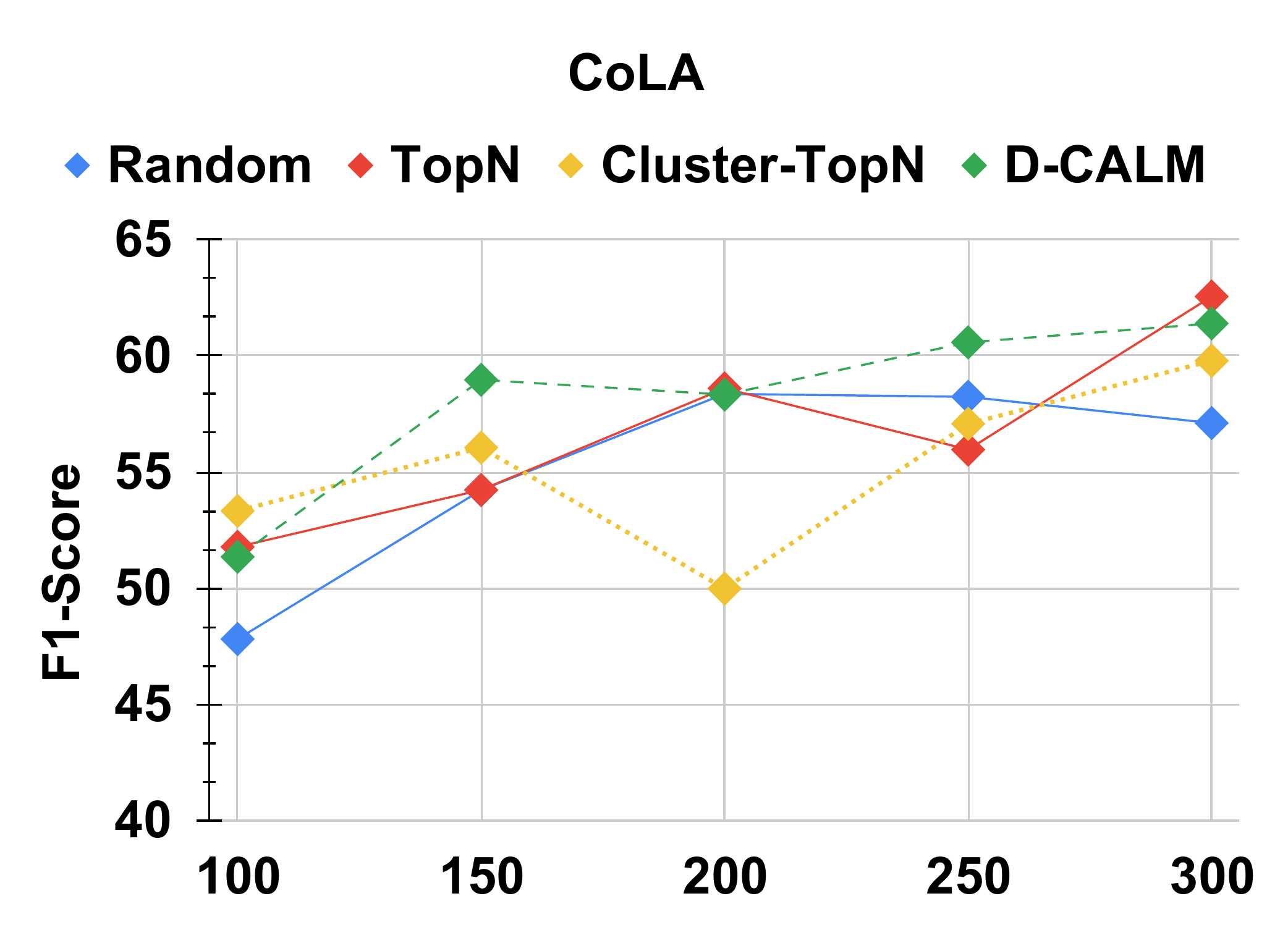}}
  \label{fig:sfig1d}
\end{subfloat}
\begin{subfloat}
  {\includegraphics[width=.316\textwidth,keepaspectratio]{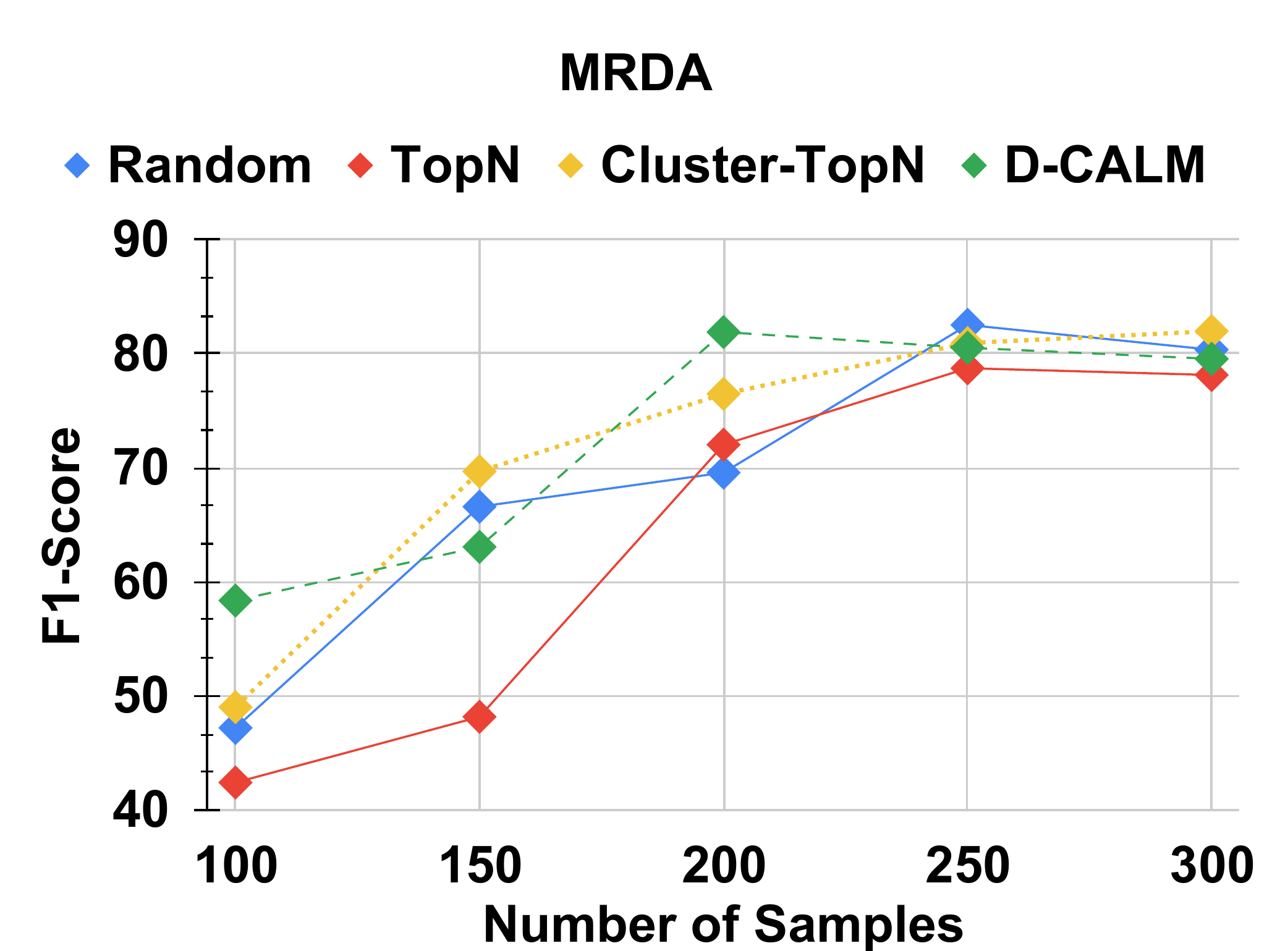}}
  \label{fig:sfig1e}
\end{subfloat}
\begin{subfloat}
  {\includegraphics[width=.316\textwidth,keepaspectratio]{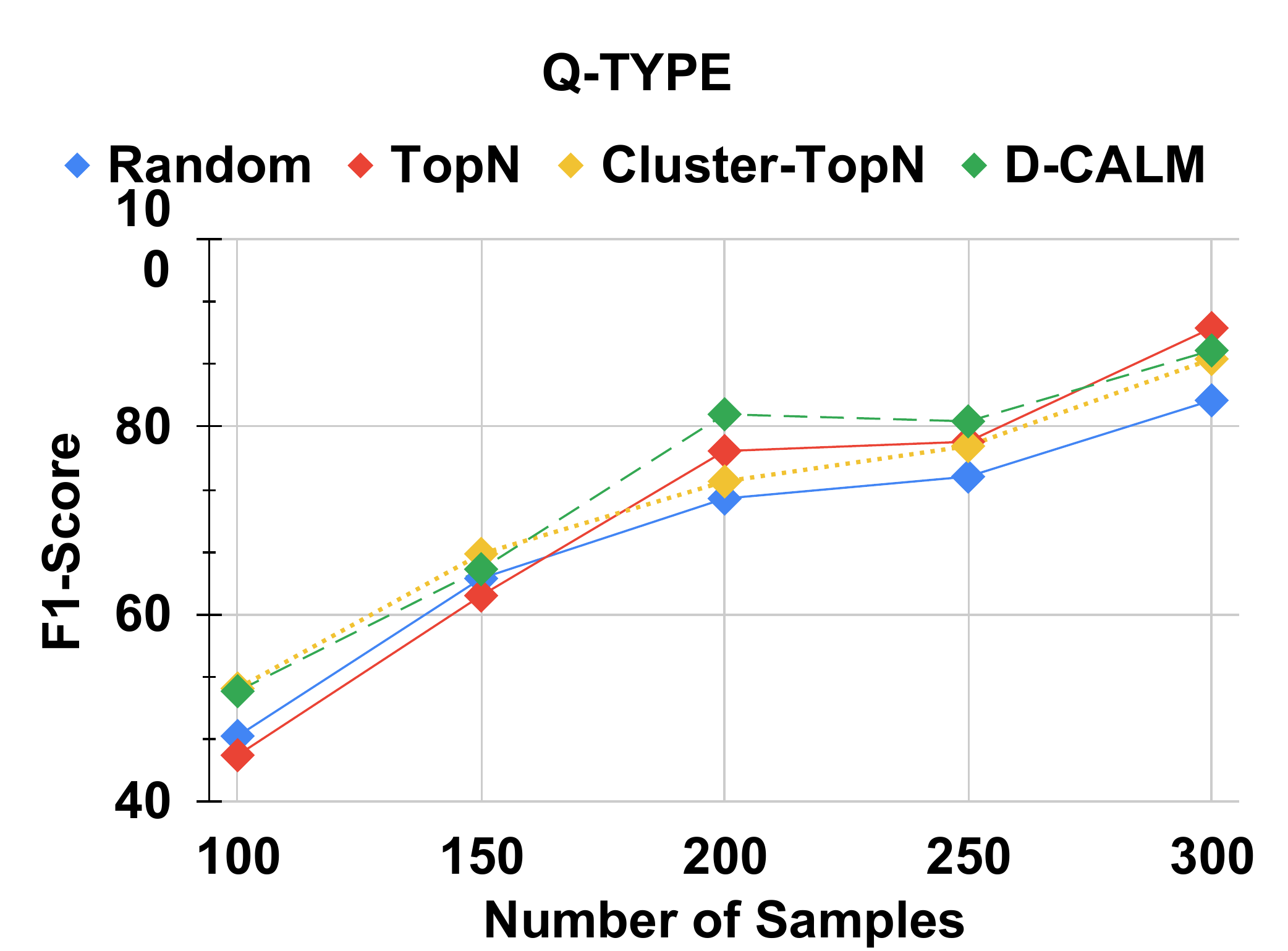}}
  \label{fig:sfig1e}
\end{subfloat}
\begin{subfloat}
  {\includegraphics[width=.316\textwidth,keepaspectratio]{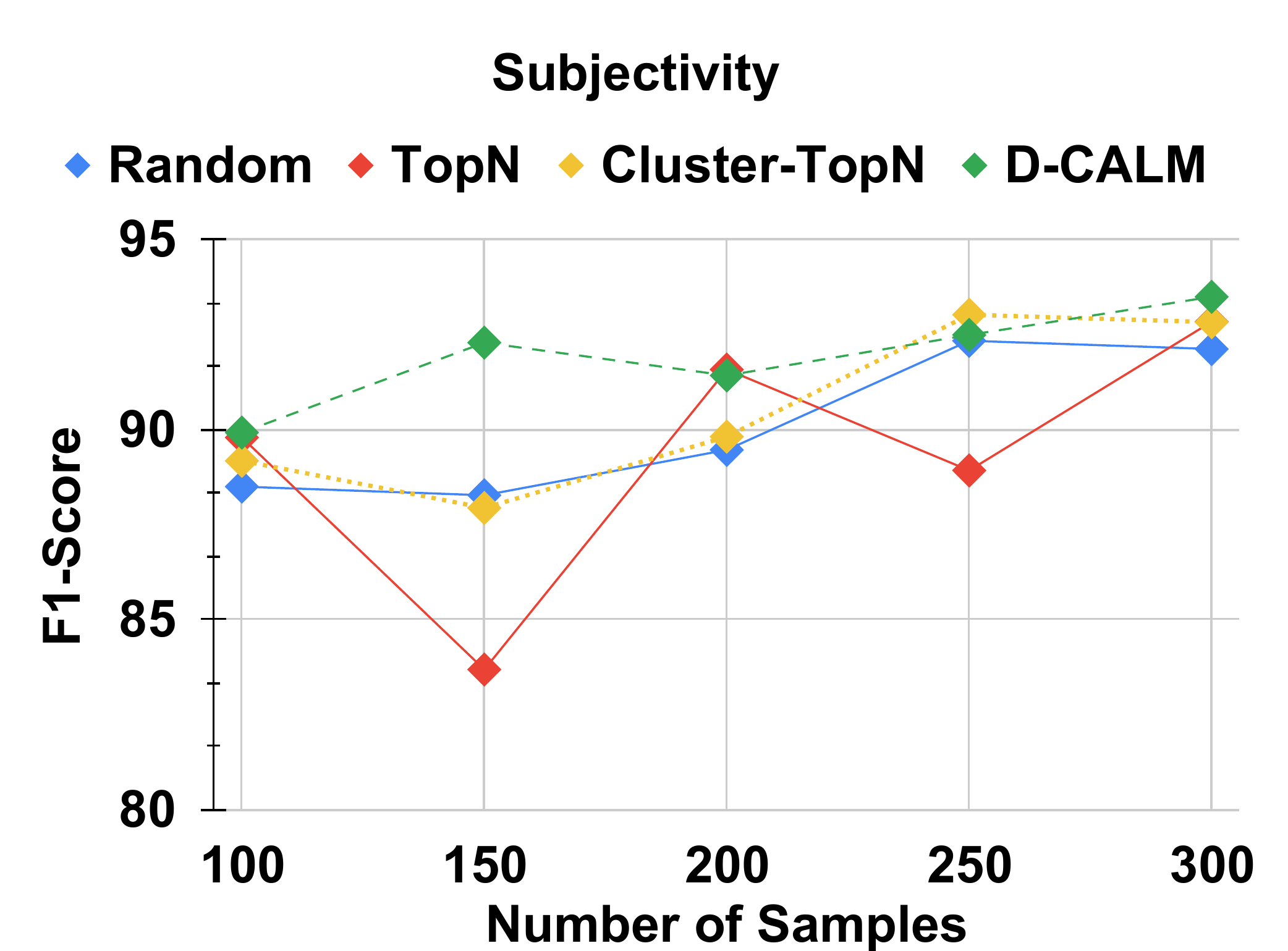}}
  \label{fig:sfig1e}
\end{subfloat}
\begin{subfloat}
  {\includegraphics[width=.316\textwidth,keepaspectratio]{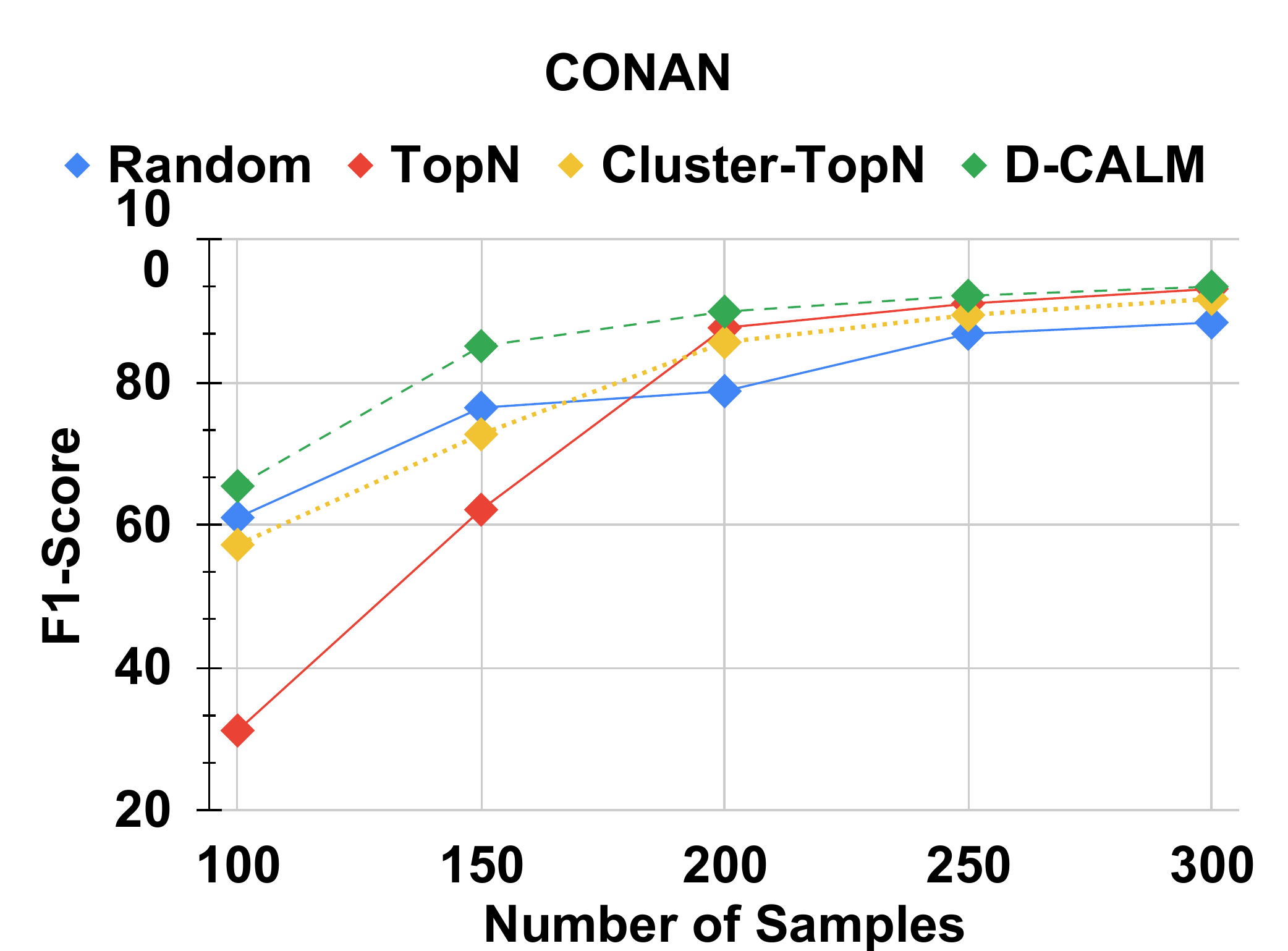}}
  \label{fig:sfig1e}
\end{subfloat}
\begin{subfloat}
  {\includegraphics[width=.316\textwidth,keepaspectratio]{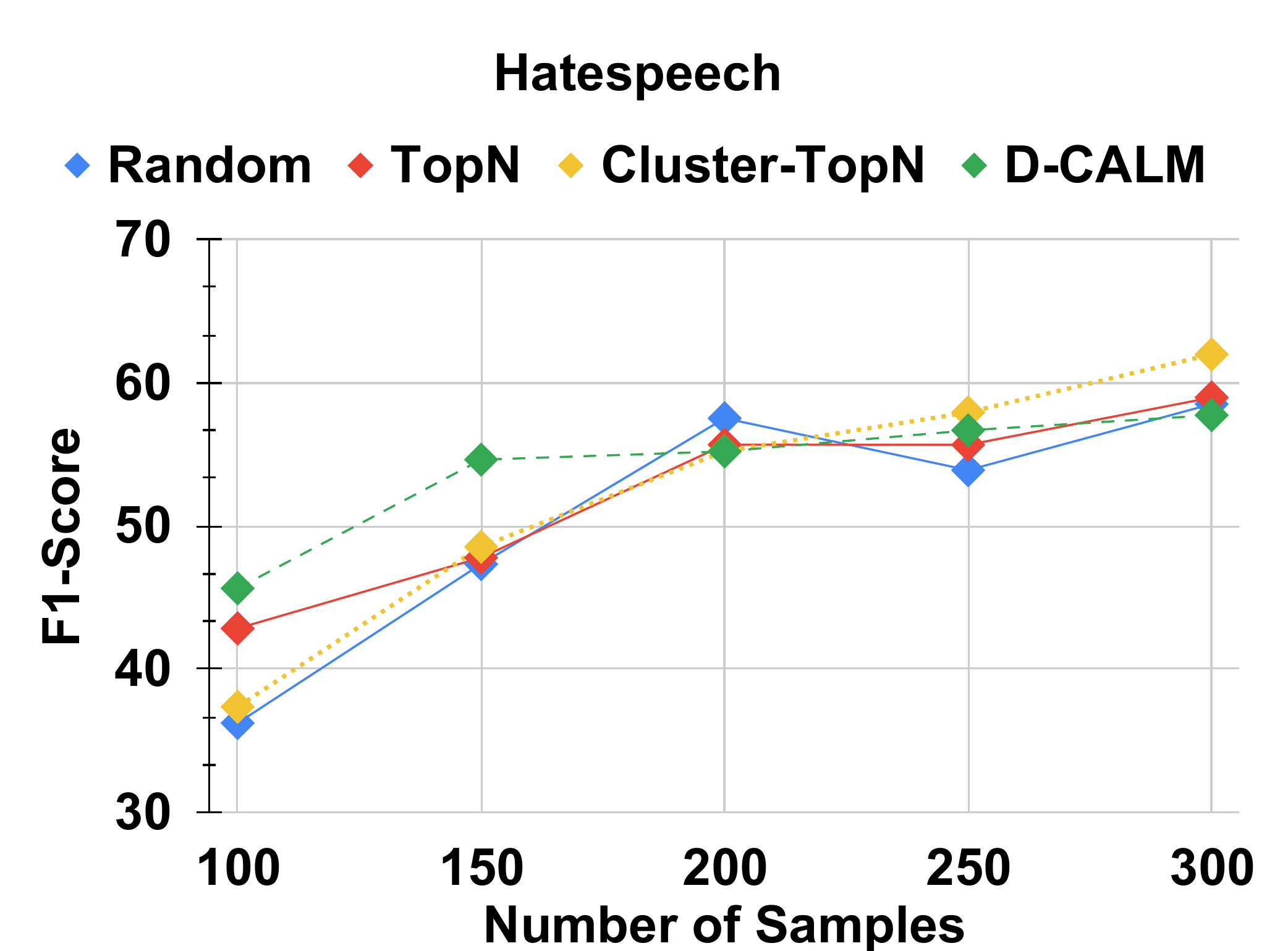}}
  \label{fig:sfig1e}
\end{subfloat}
\caption{Comparison of our proposed algorithm (\textbf{D-CALM}) and baseline approaches. \textbf{D-CALM} (green-dashed-line) consistently outperforms baseline approaches across eight datasets with Entropy as the measure of information gain and BERT as learner model.} 
\label{fig:bert-entropy}
\end{centering}
\end{figure*}

\begin{figure*}[!h]
\begin{centering}
\begin{subfloat}
  {\includegraphics[width=.316\textwidth,keepaspectratio]{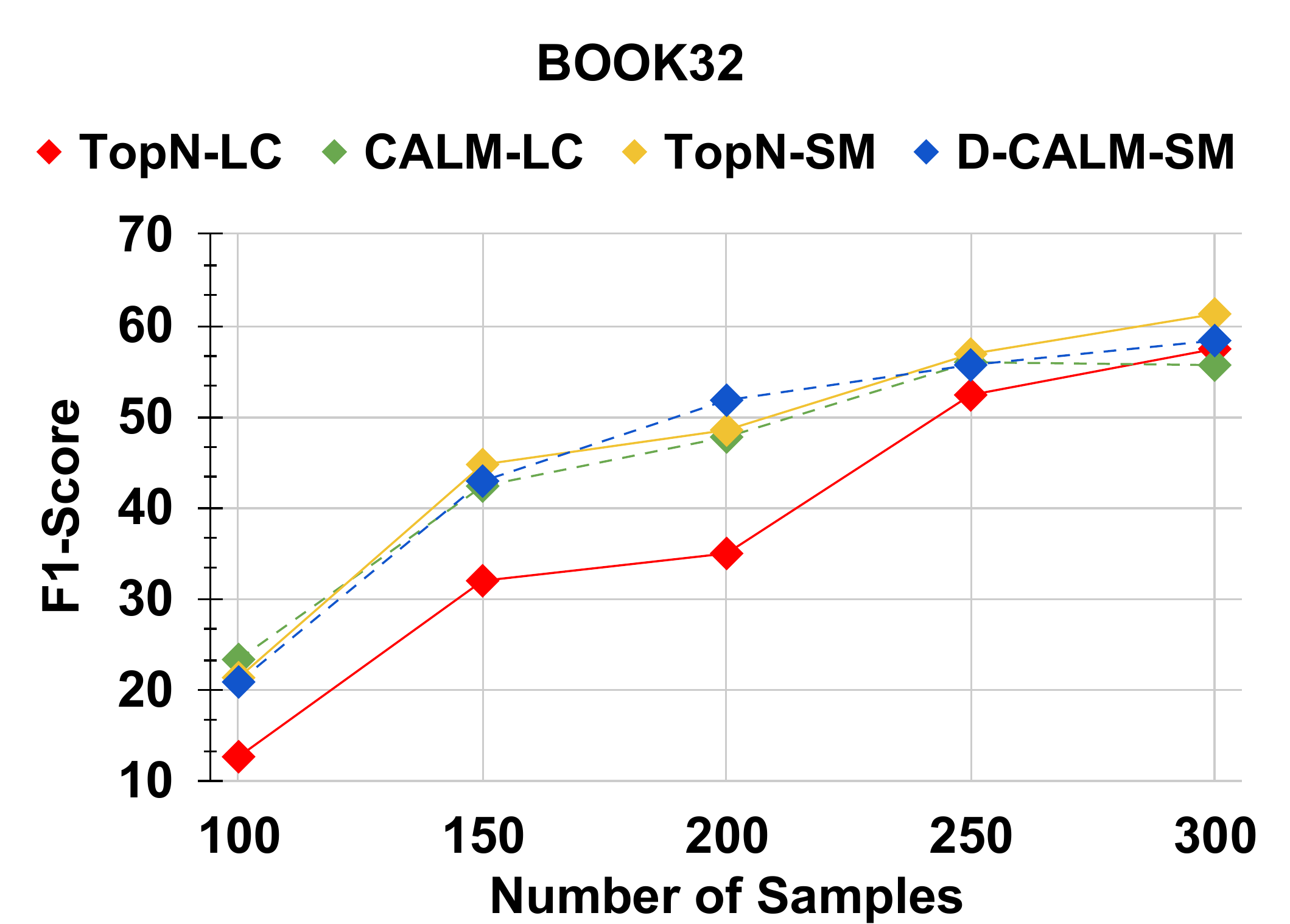}}
  \label{fig:sfig1b}
\end{subfloat}%
\begin{subfloat}
  {\includegraphics[width=.316\textwidth,,keepaspectratio]{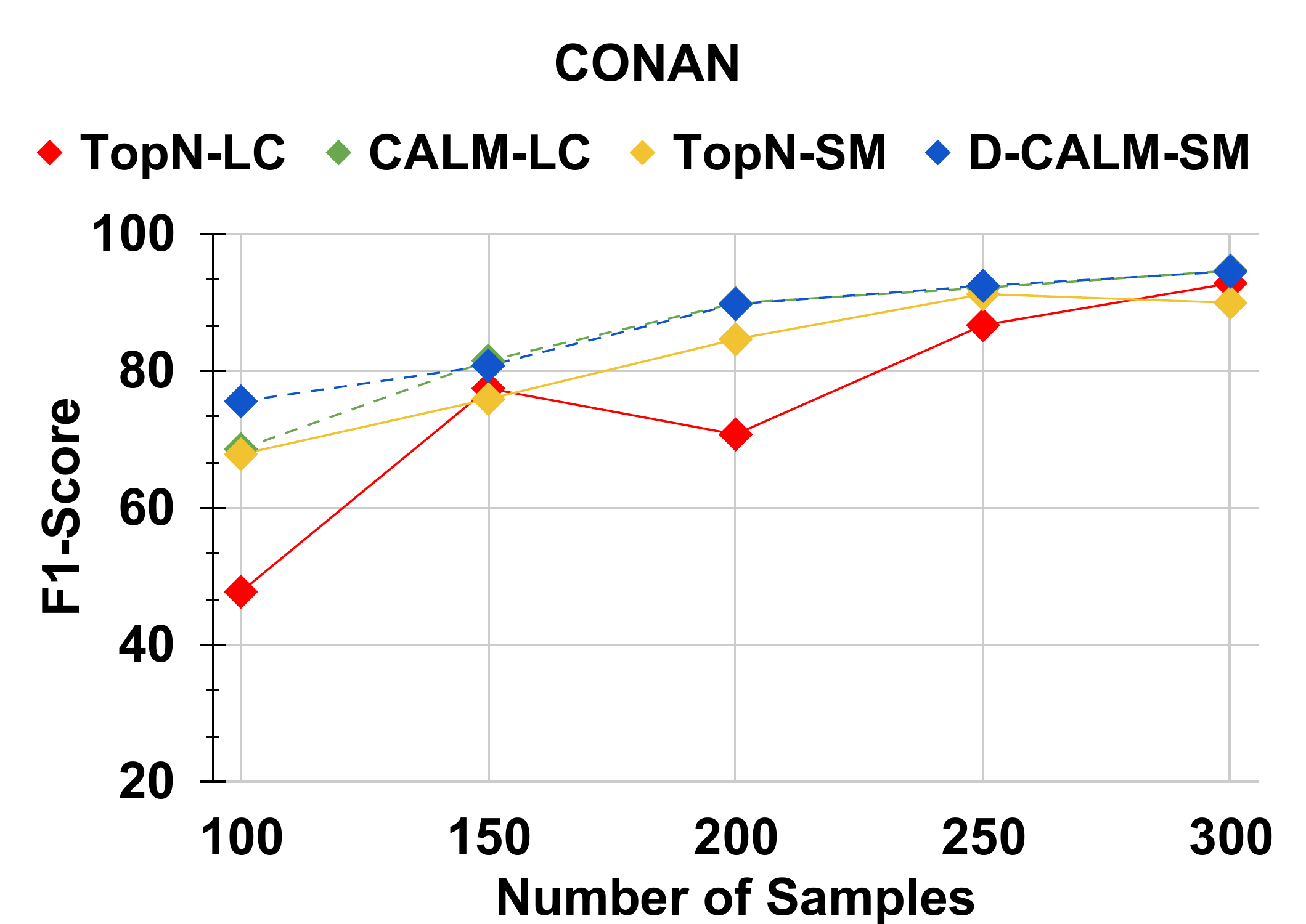}}
  \label{fig:sfig1c}
\end{subfloat}
\begin{subfloat}
  {\includegraphics[width=.316\textwidth,keepaspectratio]{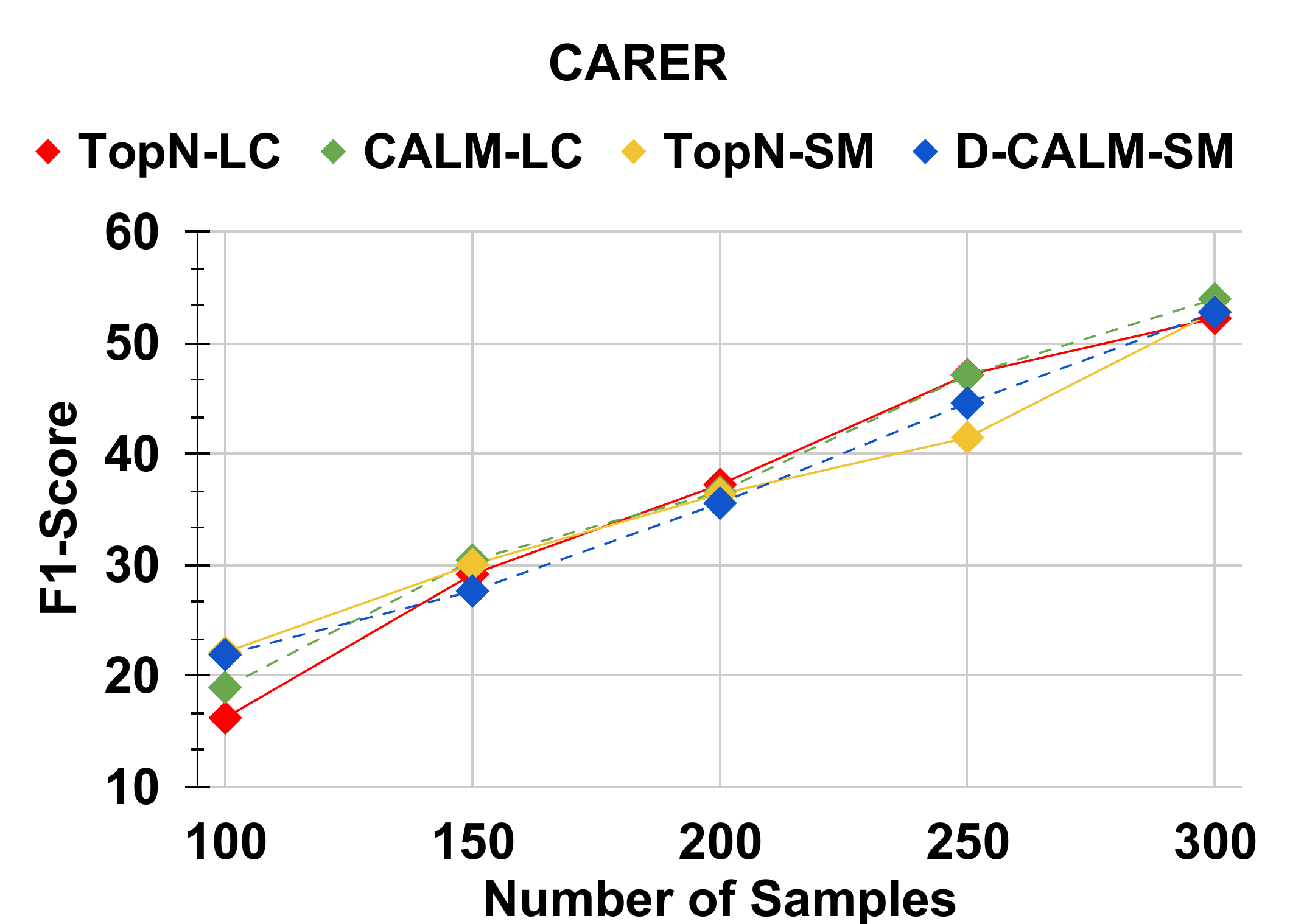}}
  \label{fig:sfig1d}
\end{subfloat}
\begin{subfloat}
  {\includegraphics[width=.316\textwidth,keepaspectratio]{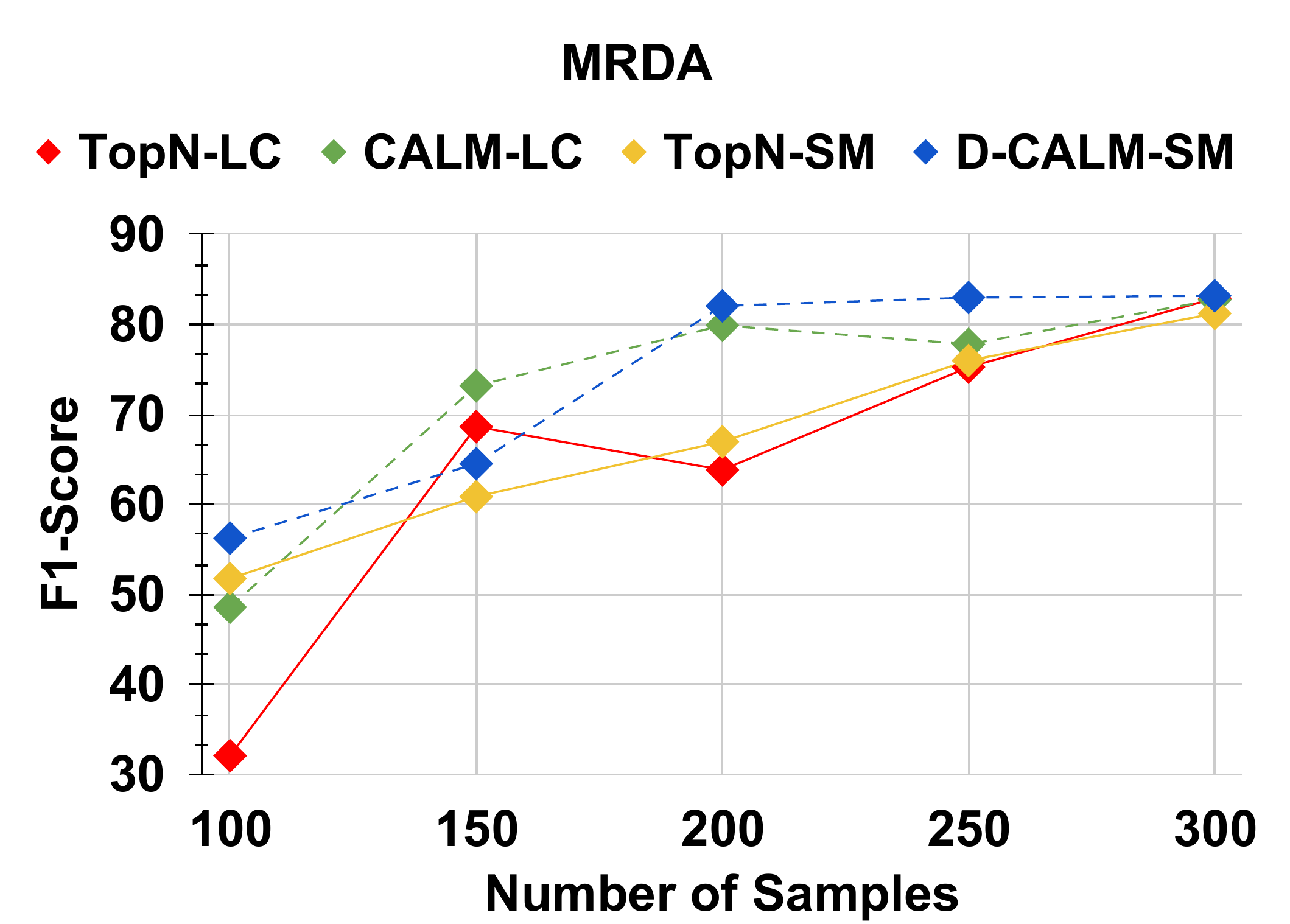}}
  \label{fig:sfig1e}
\end{subfloat}
\begin{subfloat}
  {\includegraphics[width=.316\textwidth,keepaspectratio]{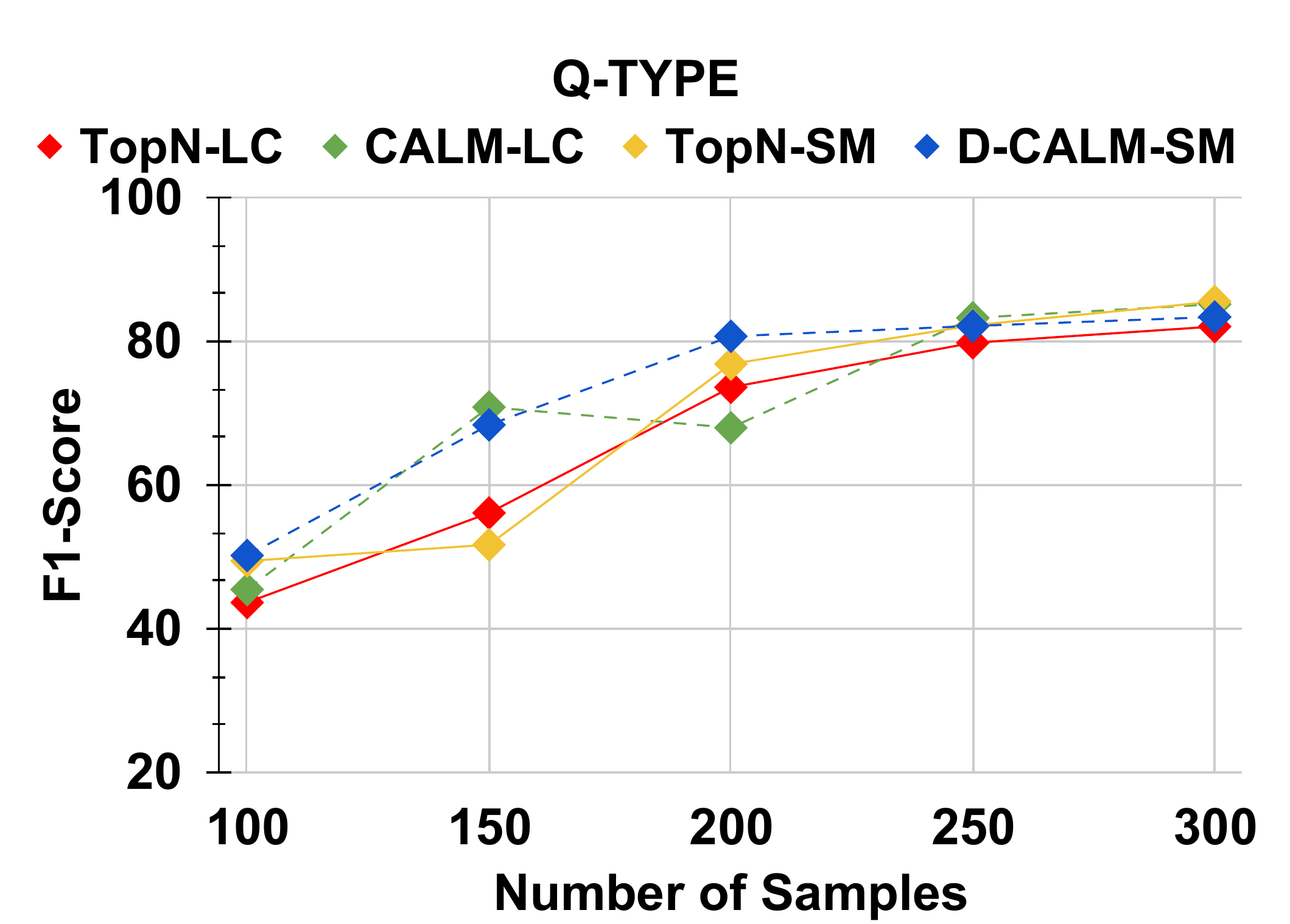}}
  \label{fig:sfig1e}
\end{subfloat}
\begin{subfloat}
  {\includegraphics[width=.316\textwidth,keepaspectratio]{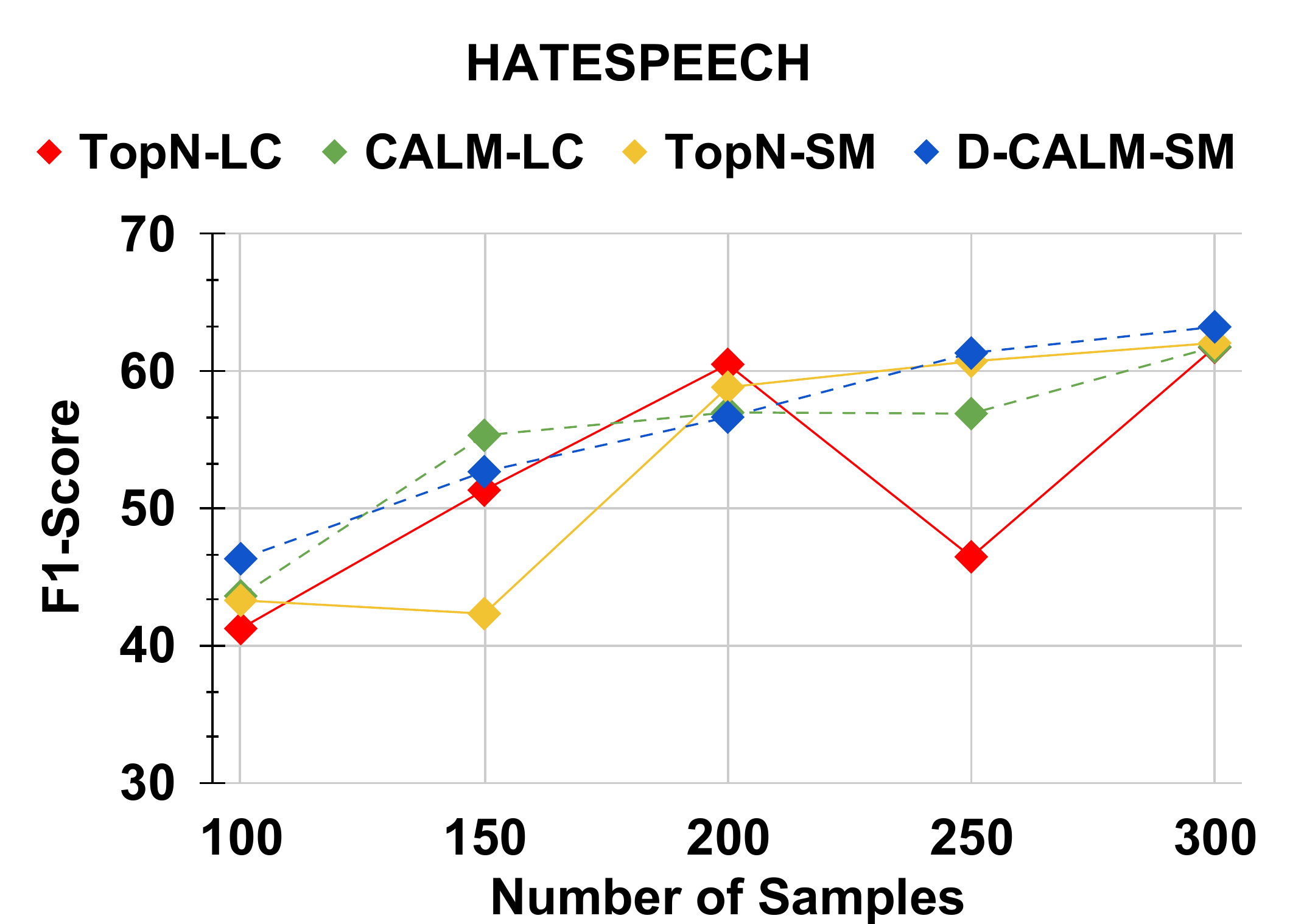}}
  \label{fig:sfig1e}
\end{subfloat}
\caption{Comparison of our algorithm with TopN approach under different measures of information gain. LC refers to Least Confident and SM refers to Smallest Margin. A consistent improvement over TopN baseline, similar to Entropy-based information gain in Figure 4 affirms the robustness of our algorithm}. 
\label{fig:bert-ig}
\end{centering}
\end{figure*}

\begin{figure*}[h]
\begin{centering}
\begin{subfloat}
  {\includegraphics[width=.318\textwidth,keepaspectratio]{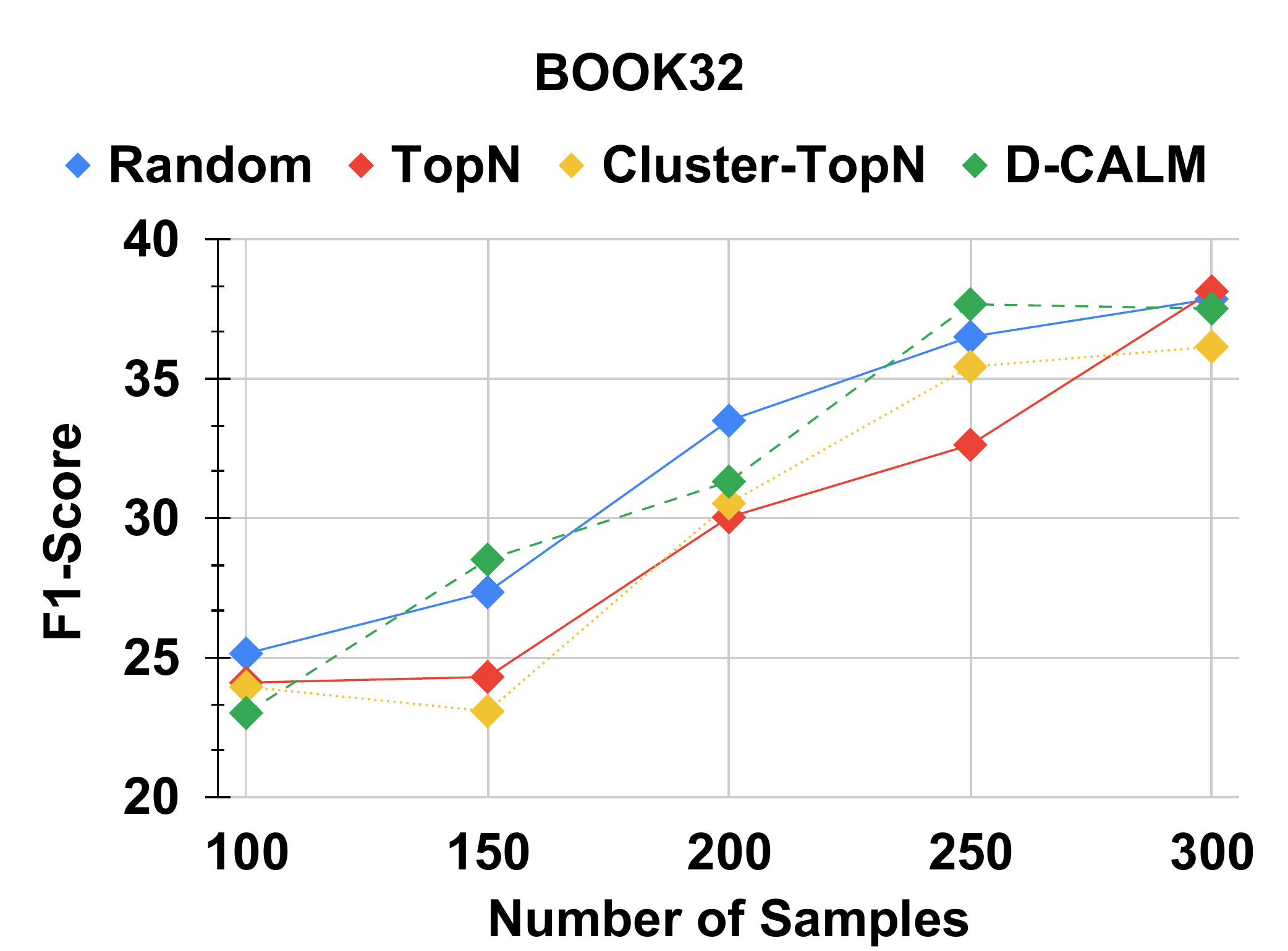}}
  \label{fig:sfig1b}
\end{subfloat}%
\begin{subfloat}
  {\includegraphics[width=.318\textwidth,,keepaspectratio]{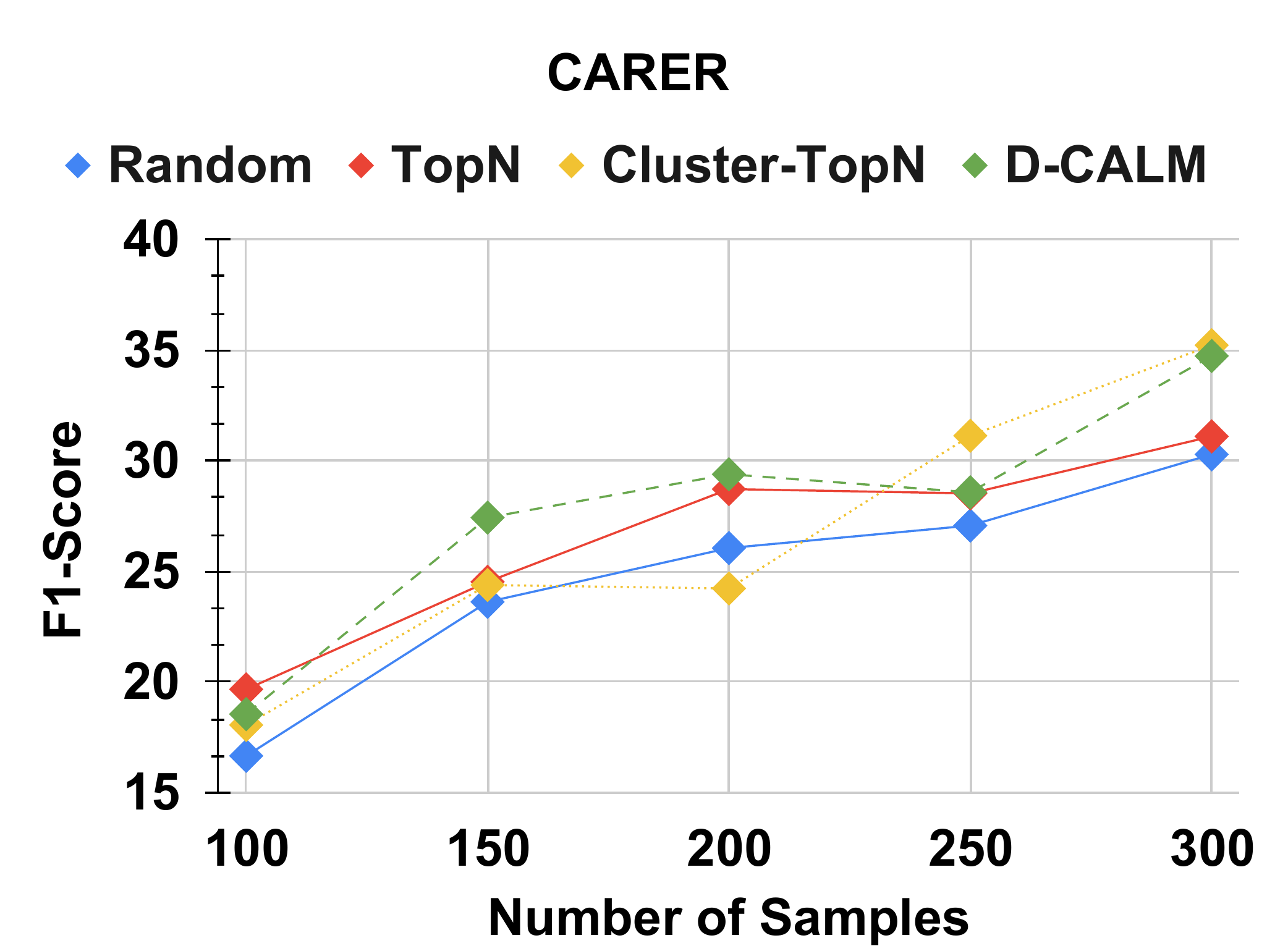}}
  \label{fig:sfig1c}
\end{subfloat}
\begin{subfloat}
  {\includegraphics[width=.32\textwidth,keepaspectratio]{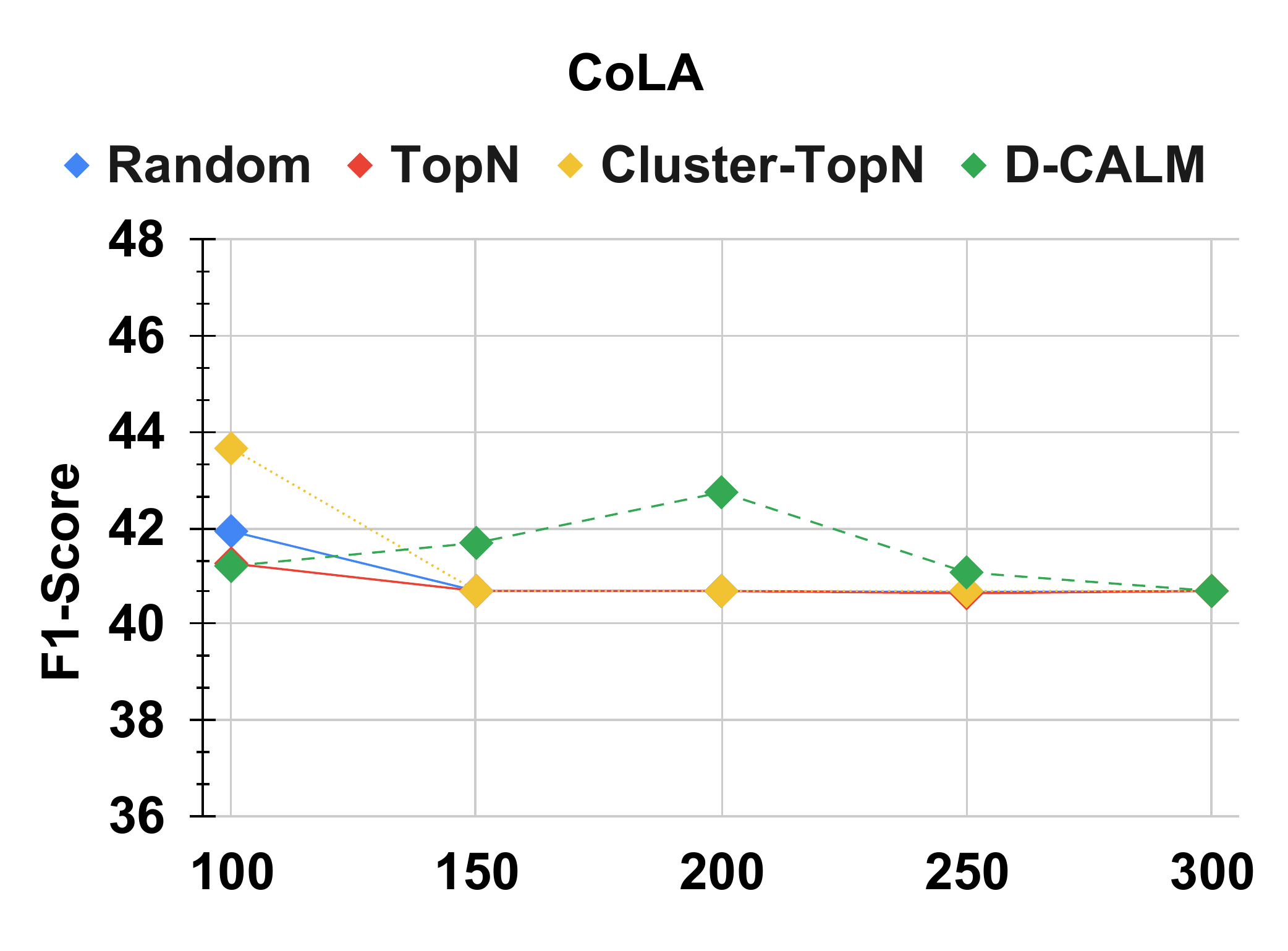}}
  \label{fig:sfig1d}
\end{subfloat}
\begin{subfloat}
  {\includegraphics[width=.32\textwidth,keepaspectratio]{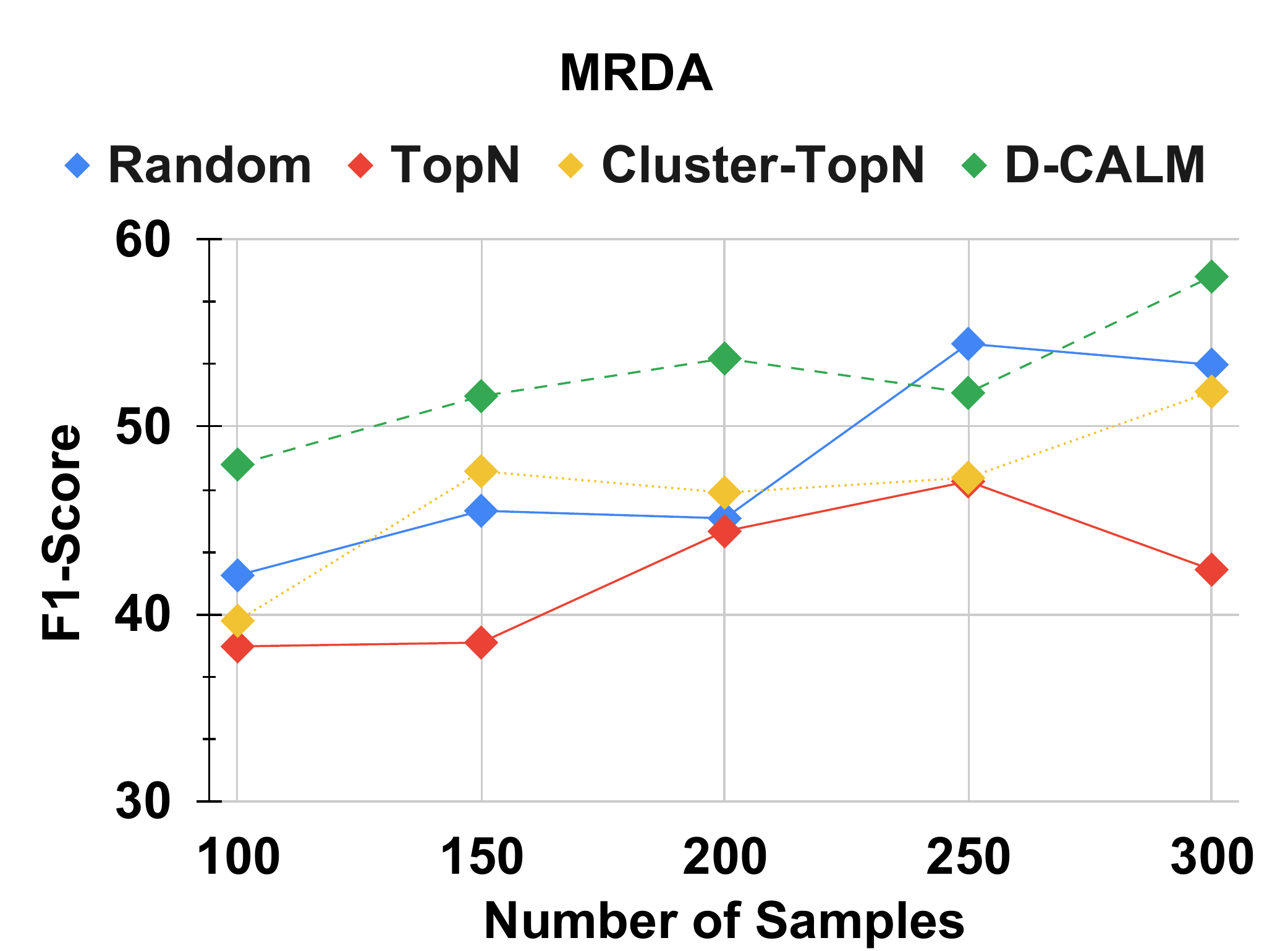}}
  \label{fig:sfig1e}
\end{subfloat}
\begin{subfloat}
  {\includegraphics[width=.32\textwidth,keepaspectratio]{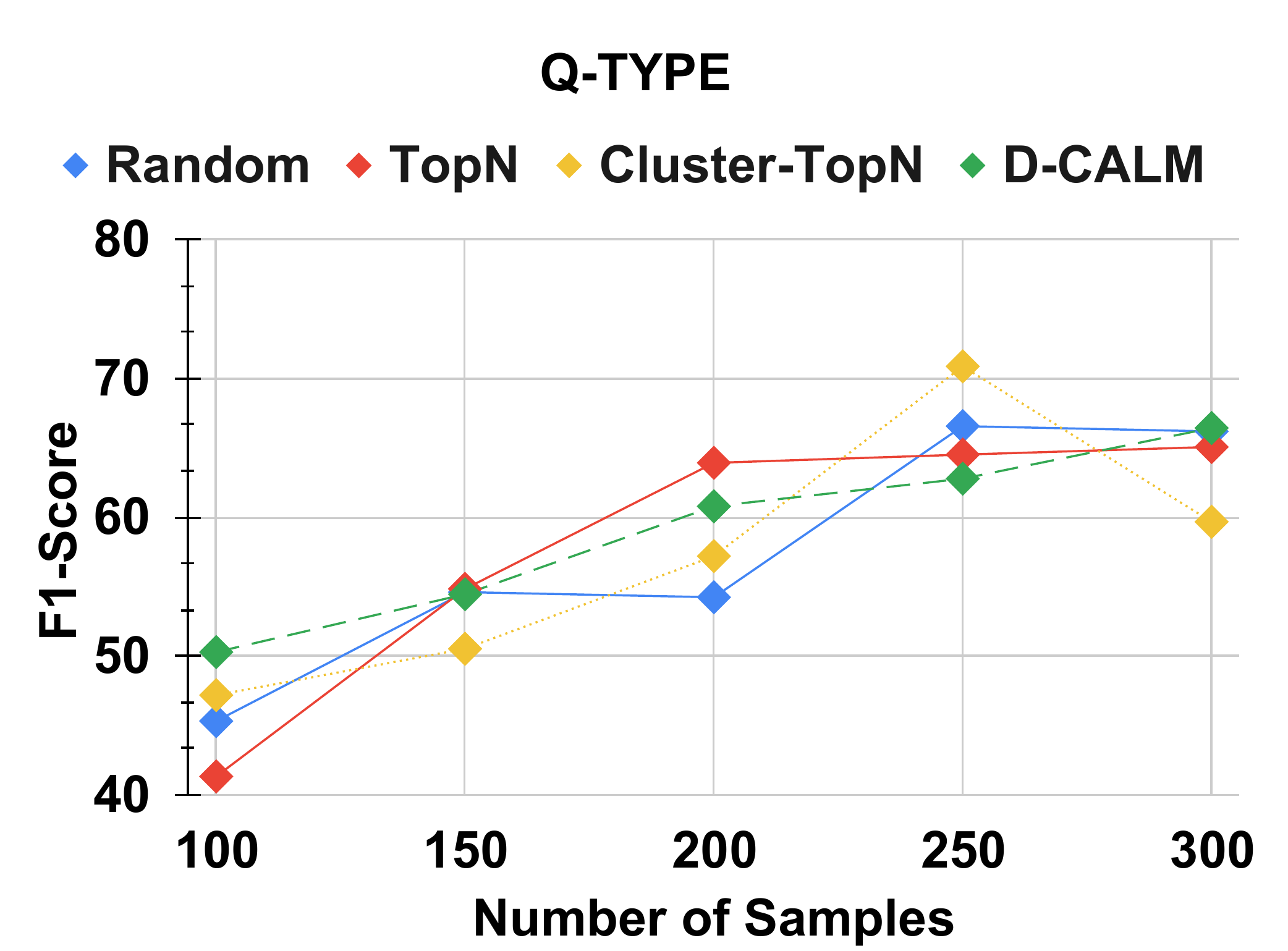}}
  \label{fig:sfig1e}
\end{subfloat}
\begin{subfloat}
  {\includegraphics[width=.32\textwidth,keepaspectratio]{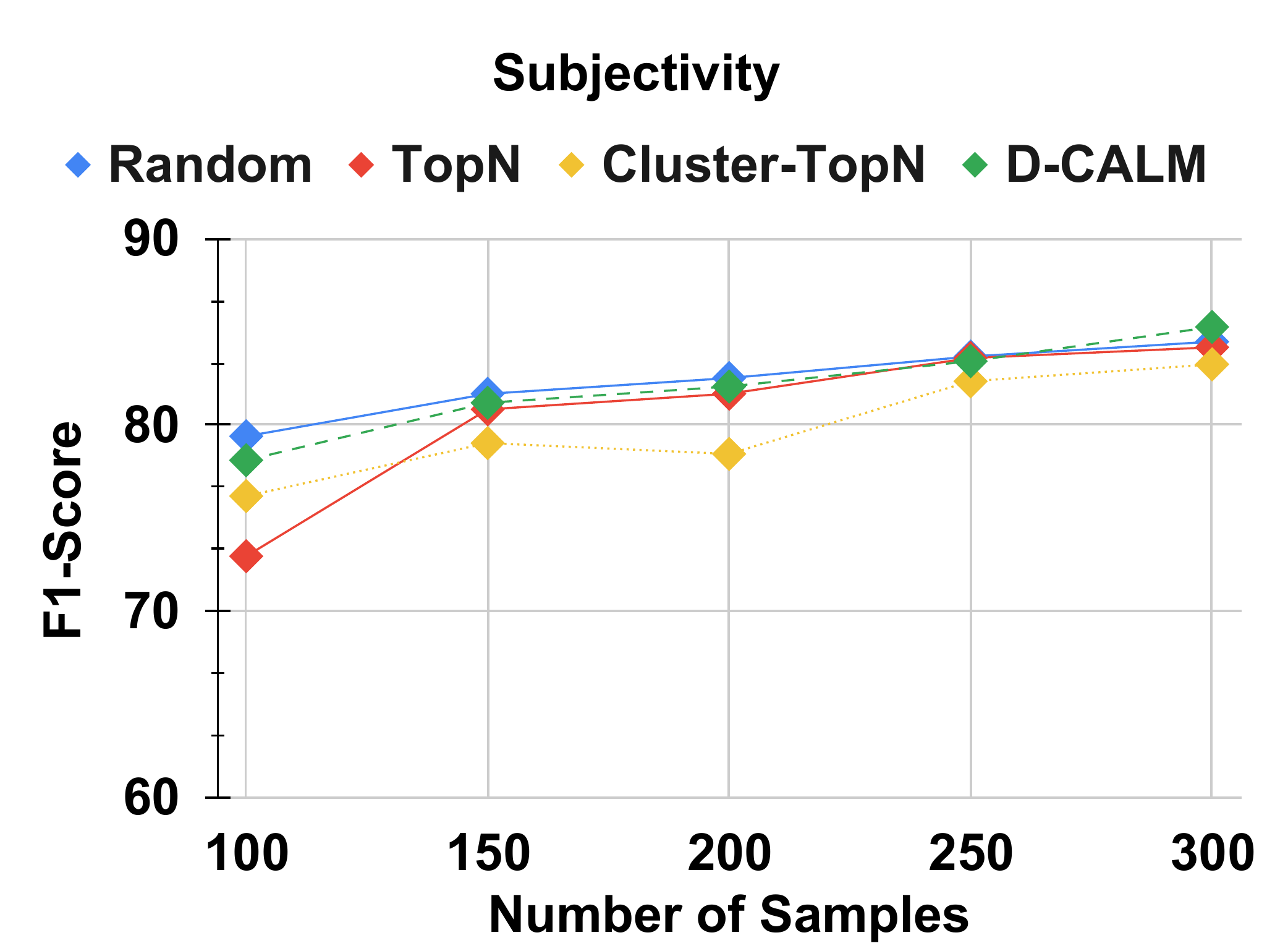}}
  \label{fig:sfig1e}
\end{subfloat}
\begin{subfloat}
  {\includegraphics[width=.32\textwidth,keepaspectratio]{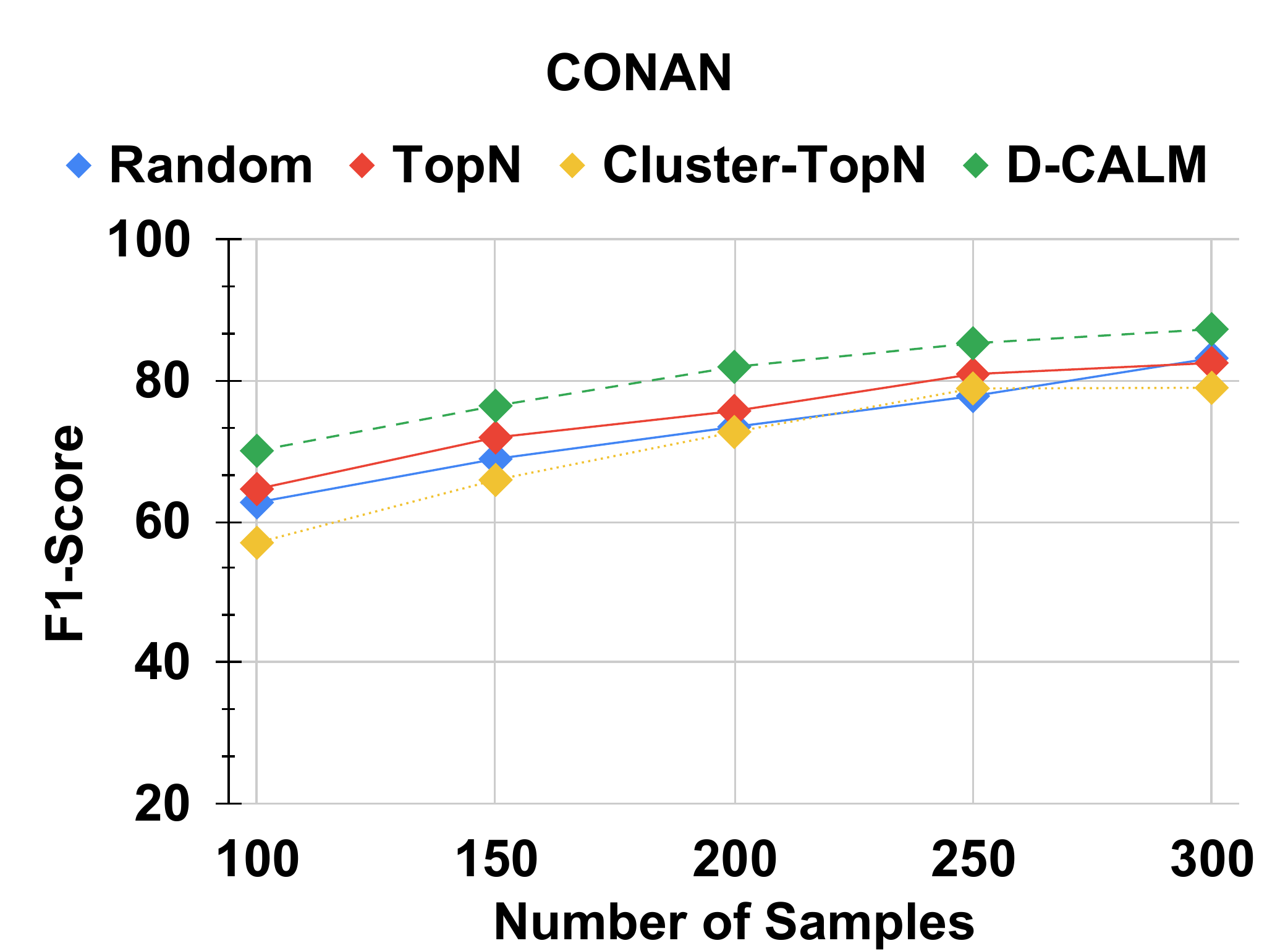}}
  \label{fig:sfig1e}
\end{subfloat}
\begin{subfloat}
  {\includegraphics[width=.32\textwidth,keepaspectratio]{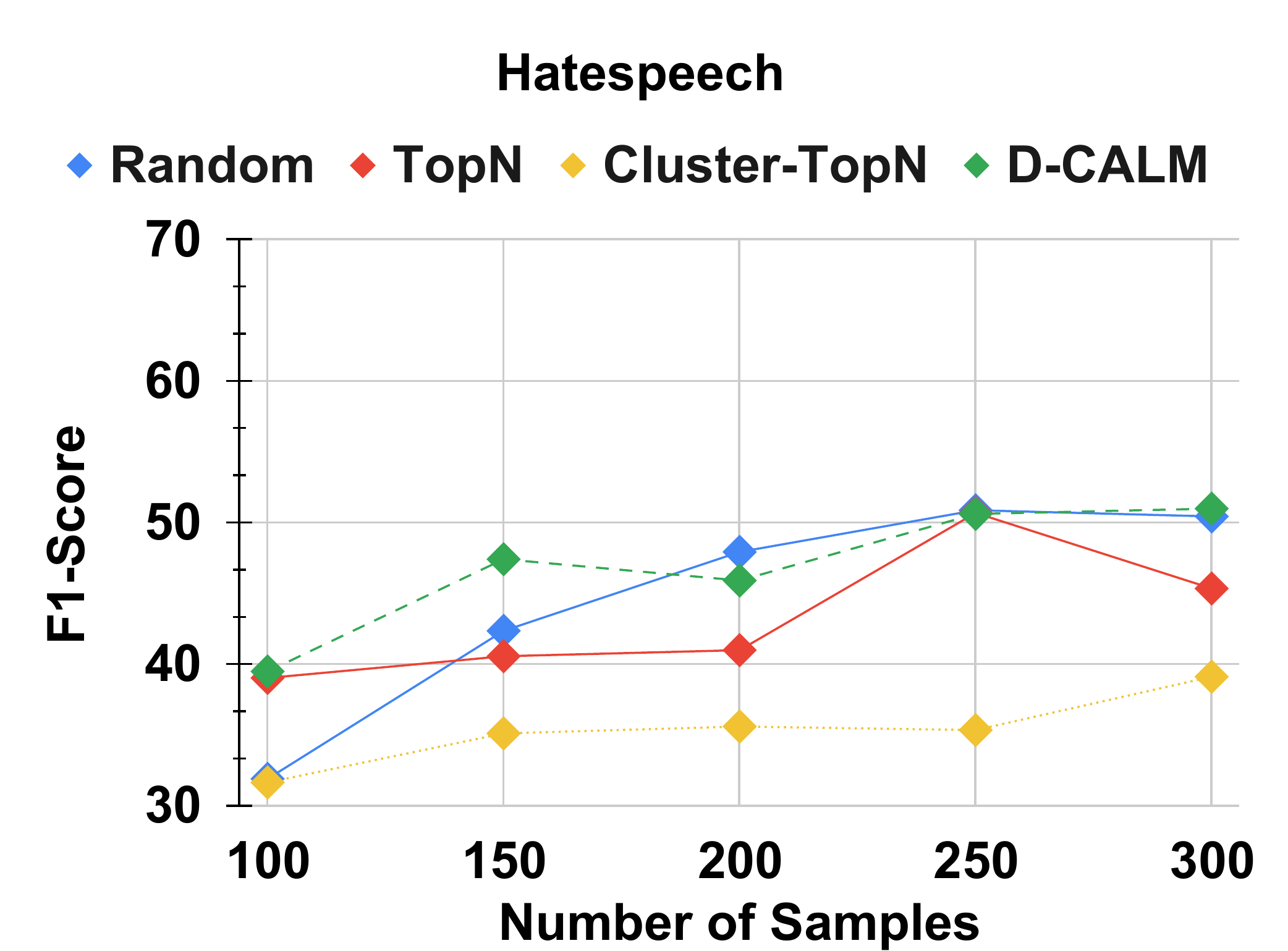}}
  \label{fig:sfig1e}
\end{subfloat}
\caption{Comparison of our proposed algorithm (\textbf{D-CALM}) and baseline approaches for SVM as learner model. Consistent improvement for SVMs in addition to BERT models (Figure 4) suggests \textbf{D-CALM} can be used for completely different types of models.} 
\label{fig:svm-entropy}
\end{centering}
\end{figure*}

\begin{figure*}[h]
\begin{centering}

\begin{subfloat}
  {\includegraphics[width=.45\textwidth,keepaspectratio]{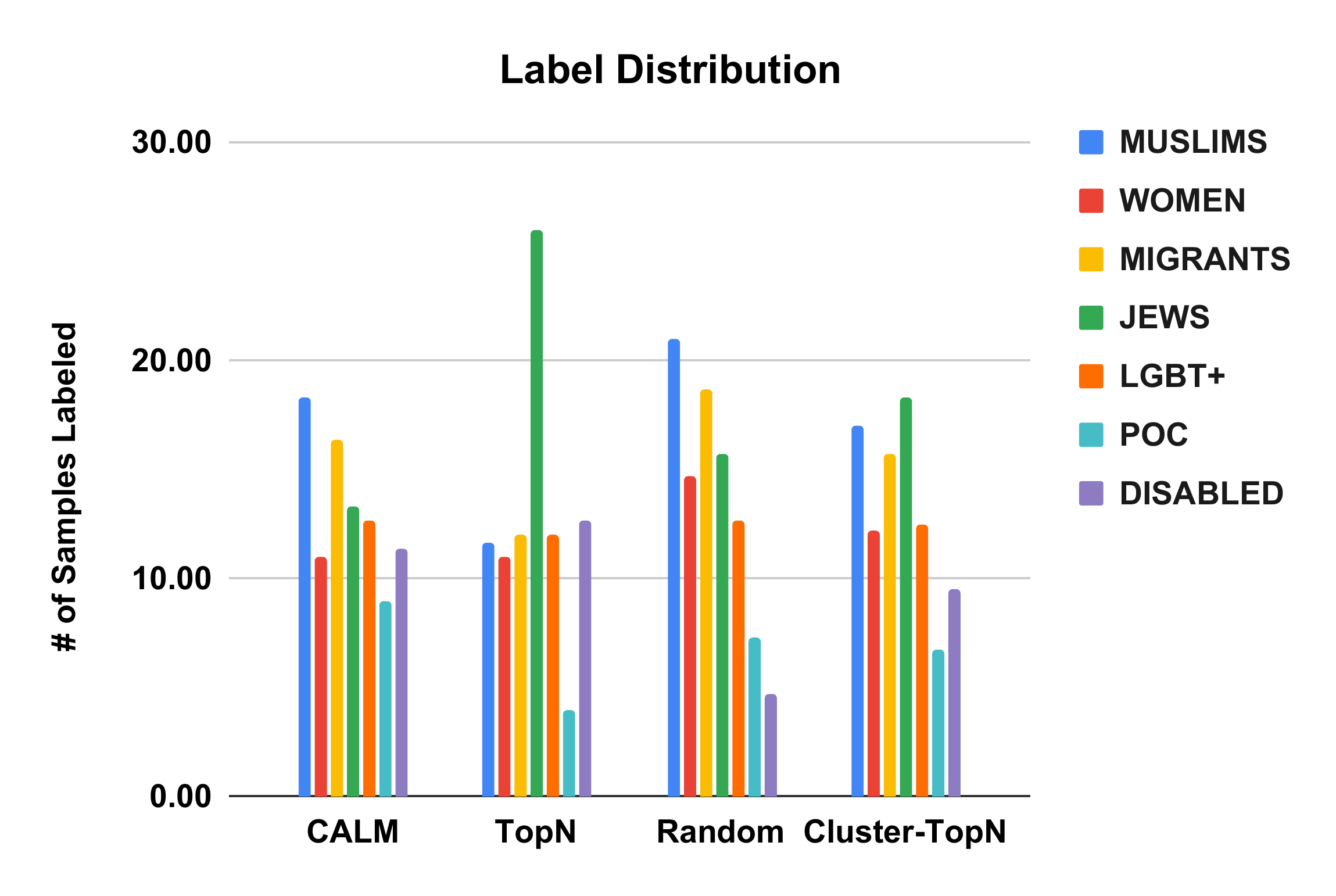}}
  \label{fig:sfig1b}
\end{subfloat}%
\begin{subfloat}
  {\includegraphics[width=.45\textwidth,keepaspectratio]{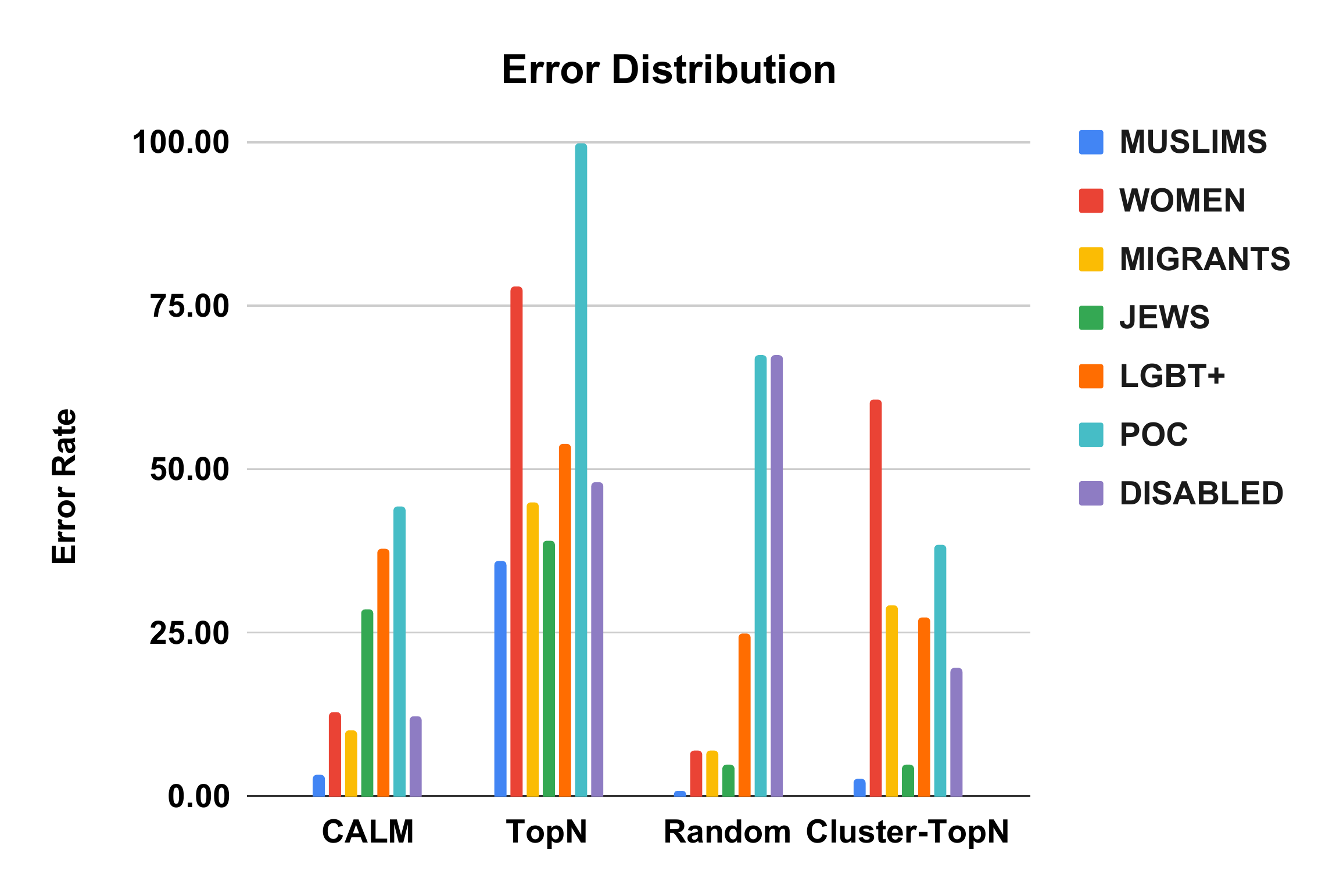}}
  \label{fig:sfig1c}
\end{subfloat}

\caption{Label and error distribution after one iteration of Active Learning on the CONAN (fine-grained hatespeech) dataset (averaged across 3 runs). \textbf{D-CALM} doesn't suffer from strong bias toward particular groups as TopN does and is more equitable toward underrepresented groups compared to random sampling.}. 
\label{fig:label-error}
\end{centering}
\end{figure*}

\section{Results and Case Study}
We first discuss findings of our experiments, followed by a case study of fine-grained hatespeech detection. Figures \ref{fig:bert-entropy},\ref{fig:bert-ig} and \ref{fig:svm-entropy} summarize the results across the eight datasets, different measures of information gain, and different models. Table \ref{tab:aggregate} summarizes relative performance across all experiments. For each experiment, we report Macro-F1 score averaged across 3 runs. We choose Macro-F1 as our metric since it provides a more holistic measure of a classifier’s performance across classes. Thus, reduction of bias is more likely to be reflected in metrics such as F1 compared to other metrics such as accuracy.
\begin{table}[]
    \centering
    \begin{tabular}{c||c|c|c}
    \hline
    \textbf{Diff.} & \multicolumn{3}{c}{\textbf{Count for BERT (IG=Entropy)}}\\
    \cline{2-4}
    \textbf{(F1)} & \textbf{DL > RND} & \textbf{DL > TN} & \textbf{DL > CTN}\\
    \hline
    > 0 & 33/40 & 32/40 & 28/40\\
    > 1 & 30/40 & 27/40 & 23/40\\
    > 3 & 24/40 & 18/40 & 16/40\\
    > 5 & 15/40 & 12/40 & 11/40\\
    > 10 & 4/40 & 4/40 & 2/40\\
    \hline
    \hline
    \cline{2-4}
    \textbf{(F1)} & \textbf{DL > RND} & \textbf{DL > TN} & \textbf{DL > CTN}\\
    \hline
    > 0 & 26/40 & 30/40 & 34/40\\
    > 1 & 21/40 & 23/40 & 30/40\\
    > 3 & 16/40 & 18/40 & 23/40\\
    > 5 & 10/40 & 11/40 & 16/40\\
    > 10 & 0/40 & 2/40 & 6/40\\
    \hline
    \end{tabular}
    \caption{Aggregated counts of \textbf{D-CALM} (DL) outperforming Random (RND), TopN (TN) and Cluster-TopN (CTN) across 8 datasets (5*8=40 data points). Diff denotes the difference of F1 score between DL and the contesting method. E.g.: diff >10 indicates the count of DL outperforming contesting methods by a difference of 10 F1 score or more.}
    \label{tab:aggregate}
\end{table}
\subsection{Experiment Results}

\paragraph{\textbf{D-CALM} consistently outperforms baselines:} From Figures \ref{fig:bert-entropy}, \ref{fig:bert-ig}, \ref{fig:svm-entropy} we can observe that \textbf{D-CALM} consistently outperforms TopN, random and cluster-TopN across all datasets. From Table \ref{tab:aggregate}, we observe that \textbf{D-CALM} beats TopN in 32 out of 40 data points for BERT, among which, the difference in F1 score is greater than 5 in 15 cases. D-CALM beats the nearest algorithm, Cluster-TopN in 28/40 (p value 0.003) for BERT and Random Sampling in 26/40 cases (p value 0.0073) for SVMs (Table \ref{tab:aggregate}). Both of these are statistically significant according to 2 population proportion test at significance level of 0.01. 

\paragraph{\textbf{D-CALM} is more robust against critical failures:} We observe from Figure \ref{fig:bert-entropy} that on occasions such as in the case of Subjectivity and MRDA, TopN can have critical failures where the model ends up with an extremely low F1 score. Although on a few occasions, we witness dips in the curves of \textbf{D-CALM}, in general, the curves are much more stable, indicating its robustness.

\paragraph{\textbf{D-CALM} is robust across different measures of information gain:} From Figure \ref{fig:bert-ig}, we see that \textbf{D-CALM} outperforms random, TopN, and cluster-TopN for different measure of information gain. Figure \ref{fig:bert-ig} does not contain the Subjectivity and CoLA because these are binary datasets and Entropy (reported in Figure \ref{fig:bert-entropy}), Least Confident and Smallest Margin become equivalent in the case of binary classification (Section \ref{framework}).

\paragraph{\textbf{D-CALM} is model-agnostic:} From Figure \ref{fig:svm-entropy}, we observe similar patterns in improvement when the learner model is SVM instead of BERT. Although the degree of improvement is smaller for SVMs compared to BERT, it is a limitation of active learning rather than \textbf{D-CALM's}, as we can see others showing smaller improvement as well. Improvement over the baselines for SVMs in addition to BERT suggests \textbf{D-CALM} is model-independent.

\paragraph{D-CALM is more robust against bias:} Since D-CALM’s focus on enforcing diversity in dynamically adjusting clusters separates it from the other methods, we can deduce that it is the bias reduction that is resulting in improved performance metrics. This is further supported by a study of label and error distribution in the following section.

\subsection{Case Study}
\label{case-study}
 We choose hatespeech detection as our case study because hatespeech and abusive content detection has been one of the most studied text classification tasks in recent years \cite{Zampieri2019PredictingTT,hassan-etal-2020-alt-semeval}. Due to its importance in many avenues, whether for AI moderation of online content \cite{ye-etal-2023-multilingual} or filtering language model data \cite{abdelali2021pretraining}, racial and gender bias in hatespeech and abusive content detection tasks have been a particular concern for the NLP community \cite{davidson-etal-2019-racial,ahn-oh-2021-mitigating}.
\begin{table*}[]
    \small
    \centering
    \begin{tabular}{l|c|c|c}
    \hline
    \textbf{Text} & \textbf{Annotated} & \textbf{Predicted} & \textbf{Error Type}\\
    \hline
    Jews are everywhere, but they are too fundamentalist. & \textit{Jews} & \textit{Muslims} & Related Class\\
    \hline
    Asians just want to talk with other asians: they are the racists... & \textit{Other} & \textit{POC} & Dataset Limitation\\
    \hline
    Migrants are just a bunch of animals, we can't accept them... & \textit{Other} & \textit{MIGRANT} & Annotation Error\\
    \hline
    
    \end{tabular}
    \caption{Examples of errors made on the CONAN dataset by our proposed approach. While our approach improves performance and mitigates bias as seen before, a better task design and annotation quality control are required to address these types of errors. \textit{Disclaimer: these examples do not reflect authors' views in any way.}}
    \label{errors}
\end{table*}
\subsubsection{Label and Error Distribution}
To understand the effect of the underlying distribution in data, we observe the label distribution of samples chosen for annotation after one round of active learning. We also observe the distribution when the same number of samples are chosen randomly. From Figure \ref{fig:label-error} we can observe that random sampling may result in small number of samples chosen for minority classes in the data such as \textit{Persons of Color} and \textit{Disabled}. The distribution of labels obtained by TopN can become particularly skewed. We see that on average, TopN samples 25+ hatespeech targeting \textit{Jews} while mostly ignoring groups such as \textit{Person of Color (POC)} with <5 samples on average. While Cluster-TopN mitigates this problem to an extent, the best results are obtained by \textbf{D-CALM}, with samples for \textit{POC} doubling compared to TopN and samples for \textit{Disabled} doubling compared to random sampling after just one iteration. The error distribution in Figure \ref{fig:label-error}, reflects the effects of this as we see errors for \textit{POC} and \textit{Disabled} are greatly reduced by \textbf{D-CALM}. It's important to note, if we had access to a large pool of labeled data, we could obtain a more balanced dataset for training. However, in a real-world scenario, before the annotation process, we only have access to unlabeled pool of data. As such, we cannot identify low-frequency classes and balance the training set. \textbf{D-CALM}, however, can obtain more samples from the underrepresented classes without knowing their true labels beforehand.
\subsubsection{Error Analysis} To understand the limitations of \textbf{D-CALM}, we manually annotated 100 errors made by the best run with BERT on the CONAN dataset after one iteration of active learning. Our key observations are listed below:
\begin{itemize}[leftmargin=*]
    \item The model can be confused on closely related classes such as \textit{Jews} and \textit{Muslims} as the hatespeech in both cases target religions.
    \item Some errors can be attributed to the limitation of annotation design. For example, CONAN contains the class Persons of Color (POC), but does not contain a separate class for racism against Asians. These instances are labeled as \textit{Other} in the data but are predicted as \textit{POC} by the model.
    \item In some cases, the error is in the original annotation, rather than the model's prediction. 
\end{itemize}
Examples of these errors are listed in Table \ref{errors}. While the first type of error can possibly be reduced with the addition of more data close to boundary regions between closely related classes, the last two types of errors need to be addressed during the design and annotation phase of the task.


\section{Conclusion and Future Work}

In this paper, we presented a novel dynamic clustering-based active learning algorithm, \textbf{D-CALM}, that can be easily adopted by the NLP community for training models with a small set of annotated data. We have shown that by focusing annotation efforts in adaptive clusters where the learner model has higher error rates, the performance can be improved substantially while reducing bias against underrepresented groups in unlabeled data. Our experiments also show that \textbf{D-CALM} is robust across different datasets, different measures of information gain, and completely different model types. In the future, our approach can be adapted for creating less biased test sets for evaluating classifiers. An exciting future direction for our approach is to adapt it for natural language generation tasks such as style-transfer \cite{atwell-etal-2022-appdia} or counterspeech generation \cite{ashida-komachi-2022-towards}.   


\section*{Limitations}
It's important to note that, in this paper, we focus on bias resulting from underlying distribution of training data. Bias that may result from pretraining of transformer models \cite{Li2021OnRA} is not within the scope of this paper. 

Although we conduct a case study of fine-grained hatespeech detection task, a collective effort from the research community is required to better quantify bias mitigation of our approach across multiple tasks and different types of bias.

Another limitation of our work is that our proposed algorithm requires dynamic adjustment of clusters. For very large datasets, this may be computationally expensive.  

\section*{Ethics Statement}
Although our proposed algorithm shows more stability and reduced bias compared to existing approaches and random sampling, it's important to observe the behavior of active learner as the algorithm may not completely eliminate bias, specifically when the annotation budget is small. This can be achieved by observing label and error variance on the evaluation data. It is also important to take into consideration the necessities of practical scenarios. In scenarios where certain type of bias is desired (e.g., higher precision), the algorithm needs to be adapted as outlined in Section \ref{d-calm-algo} 

\bibliography{anthology,custom}
\bibliographystyle{acl_natbib}




\end{document}